%% file: neurips_2026.tex
\documentclass{article}

    \PassOptionsToPackage{numbers, compress}{natbib}
 \usepackage[main, final]{neurips_2026}

\usepackage[utf8]{inputenc} 
\usepackage[T1]{fontenc}    
\usepackage{hyperref}       
\usepackage{url}            
\usepackage{booktabs}       
\usepackage{amsfonts}       
\usepackage{nicefrac}       
\usepackage{microtype}      
\usepackage{xcolor}         
\usepackage{graphicx}
\usepackage{amsmath, amsfonts}
\usepackage{subcaption}
\usepackage{wrapfig}
\usepackage{multirow}
\usepackage{pifont}

\newcommand{\cmark}{\ding{51}}%
\newcommand{\xmark}{\ding{55}}%

\title{Fusion-Free Multimodal Classification with Heterogeneous Deep Ensembles}

\author{%
  Ilya Burenko \\
  ScaDS.AI\thanks{Center for Scalable Data Analytics and Artificial Intelligence Dresden/Leipzig}, Technische Universität Dresden\\
  Dresden, Germany \\ 
\texttt{ilia.burenko@tu-dresden.de} \\
  \And
  Dmitry Vetrov \\
  Constructor University \\
  Bremen, Germany \\
}

\begin{document}

\maketitle

\begin{abstract}

In multimodal classification, late-fusion approaches classify concatenated modality-specific features extracted by unimodal neural networks. 
When modality imbalance is pronounced, various regularization techniques have been proposed to balance the learning process and overcome the inferior performance of late-fusion networks. 
In contrast, this work demonstrates that multimodal data can be effectively classified without any explicit modality fusion, using deep ensembles of unimodal networks. 
We systematically compare deep ensembles to late-fusion networks at equal parameter count and show that ensembles consistently outperform state-of-the-art late-fusion methods designed to address modality imbalance when trained from random initialization. 
This advantage also holds over intermediate-fusion techniques we evaluated and over hybrid methods that combine unimodal and multimodal classifiers. 
We propose and empirically validate a method for selecting the number of models per modality in an ensemble, avoiding computationally expensive exhaustive search. 
Under extreme modality imbalance and small ensemble sizes, the heuristic indicates that ensembles of unimodal models trained solely on the stronger modality are preferable.
As the ensemble scales up, incorporating models from the weaker modality becomes beneficial. 
Both predictions align with our empirical findings. 
However, the heuristic becomes less reliable when the diversity of ensemble members is low.
To systematically explore the challenges of optimizing multimodal models, we propose a synthetic multimodal framework that allows control over both the number of modalities and their predictive strength. 
Our findings are consistent across synthetic and real-world datasets. Finally, by fitting scaling laws to bimodal datasets, we estimate the asymptotic performance of ensembles.
\end{abstract}

\section{Introduction}
\label{sec:intro}
\input{content/section_1_introduction}

\section{Related work}
\label{sec:related_work}
\input{content/section_2_related_work}

\section{Motivation}
\label{sec:motivation}

\input{content/section_3_motivation}

\section{Synthetic multimodal dataset}
\label{sec:synthetic_data}

\input{content/section_4_synthetic_data}

\section{Deep ensembles and late-fusion networks}
\label{sec:deep_ens_and_late_fusion}

\input{content/section_5_methods}

\section{Results}
\label{sec:results}
\input{content/section_6_results}

\subsection{Selecting the number of networks in an ensemble}
\label{subsec:number_of_networks}

\input{content/section_6_2_number_of_networks}

\subsection{Results on synthetic datasets}
\label{subsec:results_synthetic}
\input{content/section_6_1_results_on_synthetic_data}
\subsection{Results on real datasets}
\label{subsec:results_on_real_data}
\input{content/section_6_3_real_data}

\subsection{Deep Ensembles versus intermediate-fusion and hybrid approaches}
\label{subsec:de_vs_interfusion_vs_i2m2}

\input{content/section_6_4_DE_vs_interfusion_vs_I2M2}

\section{Scaling laws}
\label{sec:scaling_laws}
\input{content/section_7_scaling_laws}

\section{Discussion, and future work}
\label{sec:discussion}
\input{content/section_8_discussion}

\section{Conclusion}
\label{sec:conclusion}
\input{content/section_9_conclusion}

\section{Acknowledgements and Disclosure of Funding}
\label{sec:acknowledgments}
\input{content/section_010_acknowledgments}


\bibliography{references}


\newpage
\appendix

\section{Synthetic dataset construction}
\label{sec:appendix_synthetic_dataset}
\input{content/section_appendix_synthetic_dataset}

\section{Additional results}
\label{sec:appendix_additional_results}
\input{content/section_appendix_a_additional_results}

\section{Training details}
\label{sec:appendix_training_details}
\input{content/section_appendix_b_training_details}

\section{Intermediate-fusion and hybrid approaches}
\label{sec:appendix_interfusion_and_i2m2}
\input{content/section_appendix_e_intermediate_fusion_and_i2m2}

\section{Logits-averaging vs. probability-averaging ensembles}
\label{sec:appendix_logits_vs_probs}

\input{content/section_appendix_c_logits_vs_probs}

\section{Reliability diagrams for HeDEs}
\label{sec:rel_diags_hedes}
\input{content/section_appendix_calibration}

\section{Scaling Laws}
\label{sec:appendix_scaling_laws}
\input{content/section_appendix_d_scaling_laws}

\section{Cost comparisons}
\label{sec:appendix_cost_comparisons}
\input{content/section_appendix_f_cost_comparison}


\newpage
\input{checklist.tex}

\end{document}

%% file: content/section_1_introduction.tex
In training deep neural networks for classification tasks, the primary objective is to extract discriminative features that most effectively explain the output labels. 
The success of modern unimodal neural networks is attributed to the availability of large datasets and scalable architectures \cite{udandarao_2024_no_zero_shot}, such as Transformers \cite{Dehghani_2023_scaling_vits_to_22b, dosovitskiy_2021_vit, Vaswani_2017_attention} and modality-specific models \cite{liu_2022_convnext, tan_2021_effnetv2, Wang_2023_internimage}.
These models incorporate modality-focused inductive biases, as seen in convolutional neural networks \cite{Krizhevsky_2012_imagenet_with_cnns, liu_2022_convnext}, recurrent neural networks \cite{cho_2014_rnns, Sutskever_2014_seq2seq}, and graph neural networks \cite{Monti_2016_GeometricDL}. 
Hybrid approaches further integrate multiple architectural advantages \cite{ghaleb_2024_cospeech_gestures, shi_2022_avhubert, teed_2020_raft}.

For multimodal classification, classifiers use multiple input modalities, which may originate from heterogeneous data sources \cite{liang_2024_foundations_survey_on_mm_ml}. 
Combining intermediate features extracted from given modalities can make the output feature vector more informative than in unimodal scenarios. 
An extensive body of work has studied fusion techniques, from early- and intermediate-fusion to late-fusion approaches \cite{fukui_2016_bilinear_pooling, nagrani_2021_mid_fusion, wang_2023_beit3, zadeh_2017_tensor_fusion, zhang_2023_dynamic_fusion_depth_data, zhang_2024_mla}.
We refer to \cite{liang_2024_foundations_survey_on_mm_ml} for a comprehensive survey. 
Typically, the existing fusion approaches for multimodal classification use unimodal neural networks that perform well on the given modalities and optimize the interplay between the unimodal networks.

Regardless of the type of fusion, access to multiple modalities should improve classification. 
However, research has demonstrated that naive late-fusion networks, which classify based on concatenated features from unimodal networks, can yield lower accuracy compared to the best-performing unimodal classifier \cite{wang_2020_what_make_training_hard}. 
To address this unexpected outcome, recent studies have examined the problem from both theoretical \cite{huang_2021_what, lu_2023_theory_mm_learning, yedi_2024_understanding_unimodal_bias} and practical perspectives \cite{peng_2022_ogm}, seeking optimal methods for combining unimodal features in multimodal classification. Although some progress has been achieved in this respect \cite{jiang_2025_aug, kontras_2025_mcr}, it is still an open question how to optimally combine unimodal features in late-fusion multimodal classification.

In this work, we study how ensembles of unimodal models trained on available modalities perform on multimodal classification tasks. We refer to such ensembles as heterogeneous deep ensembles (HeDEs) when an ensemble comprises unimodal models trained on different modalities, contrasting them with homogeneous deep ensembles (HoDEs), which consist of models trained on a single modality. Notably, neither HeDEs nor HoDEs explicitly perform any form of modality fusion.


Our main contributions are as follows:
\begin{enumerate}
\item 
We systematically compare deep ensembles to late-fusion networks at equal parameter count and show that ensembles consistently outperform state-of-the-art late-fusion methods specifically designed to address modality imbalance. 
This advantage also holds over the intermediate-fusion techniques we evaluated, for both CNN and Transformer architectures, as well as over hybrid approaches that combine unimodal and multimodal predictions.
\item 
We propose and empirically validate a heuristic for selecting the number of models per modality in an ensemble, avoiding computationally expensive exhaustive search. 
Under extreme modality imbalance, the heuristic favors homogeneous ensembles at small sizes but predicts a transition to heterogeneous ensembles as the ensemble grows, which is consistently confirmed by our experiments.
\item 
We propose a synthetic multimodal dataset construction that uses real-world data and provides control over the number of modalities and their predictive strength. 
Unlike prior synthetic benchmarks, our construction allows training overparameterized networks, revealing optimization challenges observed in practice. 
Our findings are consistent across synthetic and real-world datasets of varying sizes and input modalities.
\item 
We fit scaling laws to both synthetic and real-world bimodal datasets and show that the test losses of unimodal models reliably predict the asymptotic test loss of the heterogeneous deep ensemble.
\end{enumerate}

%% file: content/section_2_related_work.tex
\paragraph{Late-fusion multimodal networks.}In multimodal classification tasks, discrepancies in generalization rates across modalities often lead to inferior performance of a late-fusion multimodal network compared to the strongest unimodal baseline \cite{wang_2020_what_make_training_hard, greedy_nature_wu_2022}.
To address this, several methods have been developed to promote balanced learning across modalities.
For example, certain studies modulate gradient vectors to suppress information from dominant modalities and enhance learning from less-represented modalities \cite{li_2023_agm, peng_2022_ogm, greedy_nature_wu_2022}.
A related approach involves incorporating regularization terms into the loss function to alter gradient directions and magnitudes, as demonstrated in \cite{huang_2025_inforeg, yang_2024_lfm}.
Building on this, a combination of gradient modulation and loss regularization has also been explored in \cite{wei_2024_mmpareto, zhang_2023_dynamic_fusion_depth_data}.
Another line of research focuses on updating modality-specific weights independently of the weights used to extract features from other modalities \cite{hua_2024_reconboost, zhang_2024_mla}.
Additionally, some work uses unimodal pre-trained features \cite{du_2023_umt_and_ume} or updates them iteratively \cite{fan_2023_pmr} to steer the multimodal learning process.
Beyond these, a boosting-based approach was proposed in \cite{jiang_2025_aug}, while a game-theoretic framework for multimodal classification was examined in \cite{kontras_2025_mcr}.

\paragraph{Deep Ensembles.}In classification tasks, deep ensembles aggregate the output probabilities of individual deep neural networks, resulting in improved performance, better uncertainty estimation, and enhanced robustness to dataset shift \cite{lakshminarayanan_2017_deep_ensembles}.
Ensembles combine features learned by individual models \cite{allen_2020_towards_understanding_ensemble} and tend to generalize more effectively when base learners, initialized differently, converge to distinct regions of the loss landscape \cite{fort_2019_deep_ensembles_landscape, sadrtdinov_2023_to_stay_or_not_to_stay}.
The term "heterogeneous deep ensembles" was coined in previous work \cite{abe_2022_are_deep_ensembles_necessary} but for the ensembles of models of different architectures rather than ensembles of models trained on different modalities.
Optimal ensemble performance requires training deep neural networks independently on the complete dataset. 
Performance degrades when base learners are optimized with a joint loss \cite{jeffares_2023_joint_training_learner_collision} or when subsets of the data are used for individual networks \cite{nixon_2020_bootstrapped_not_for_deep_ensembles}.
Although increasing ensemble size is straightforward, the resulting performance is often comparable to that of a single, larger network of similar total capacity \cite{abe_2022_are_deep_ensembles_necessary, dern_2025_theoretical_limitations_of_ensembles}.
Furthermore, the asymptotic behavior of loss functions has been analyzed for ensembles of unimodal text \cite{kim_2025_scaling_laws_for_text_models} and vision \cite{lobacheva_2020_power_law_in_des} models.

A natural question is whether the benefits of independent training extend to the multimodal setting, where models can be trained independently on different modalities. 
Makino et al.~\cite{makino_2023_incidental_corr} showed that in extremely low-data regimes, ensembles of two 2-layer MLPs trained on different modalities outperform their multimodal counterpart. 
The work closest to ours, I2M2 \cite{madaan_2024_i2m2}, demonstrates that a hybrid approach averaging log-probabilities of unimodal and multimodal networks outperforms its individual components. 
However, neither of these works systematically compares deep ensembles against imbalance-aware late-fusion methods, and fair parameter count comparisons are limited \cite{madaan_2024_i2m2}. 
Our work fills this gap by systematically comparing deep ensembles against these approaches under fair parameter count across a range of modality imbalance levels.

\paragraph{Synthetic multimodal datasets.}
Previous works have proposed synthetic multimodal datasets to study interactions between modalities in controlled settings \cite{kontras_2025_mcr, liang_2023_information_decomposition, liang_2023_factorized_contrastive_learning}. 
These approaches construct synthetic data by sampling shared and modality-specific vectors from multivariate normal distributions and processing them through shallow MLPs.
While these approaches allow control over mutual information between modalities, they do not involve training overparameterized networks on real-world inputs and cannot reproduce optimization challenges observed in practice, such as the faster convergence of the weaker modality \cite{wang_2020_what_make_training_hard}. 
In this work, we propose a synthetic multimodal dataset construction that addresses these limitations.

%% file: content/section_3_motivation.tex
We refer to the modality with lower validation performance as the
\textit{weak} modality and the one with higher validation accuracy as the
\textit{strong} modality.
Under significant modality imbalance, naive late-fusion networks
underexplore the slower-converging modality, which is typically the
\begin{wrapfigure}{r}{0.4\textwidth}
    \centering
    \includegraphics[width=\linewidth]{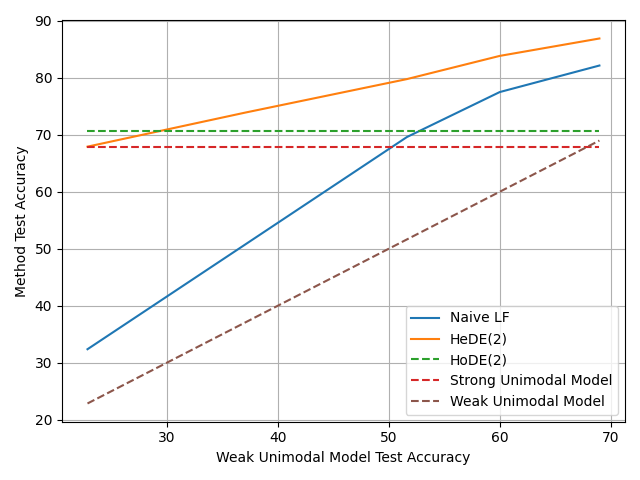}
    \caption{Test accuracy on bimodal synthetic CIFAR-100 as the
    weaker modality's share increases from 5\% to 50\% (left to right),
    with the stronger modality fixed at 50\%.}
    \label{fig:motivation_picture}
\end{wrapfigure}
stronger one, leading to worse performance than that of the best
unimodal baseline~\cite{peng_2022_ogm, wang_2020_what_make_training_hard}.
When modalities are more balanced, both subnetworks have comparable
time to extract discriminative features, and the performance of the
late-fusion network improves.
An alternative regime, where the weak modality converges more slowly
(e.g., due to noise corruption), is studied in
Sec.~\ref{subsec:appendex_noise_corruption}.
In this section, we demonstrate that deep ensembles of unimodal
networks outperform late-fusion classifiers across a range of modality
imbalance levels.

To illustrate this, we use the synthetic construction from
Sec.~\ref{sec:synthetic_data} on CIFAR-100~\cite{krizhevsky_2009_cifar100}.
We fix the data available to the stronger modality at 50\% of the
original dataset and steadily increase the weaker modality's share
from 5\% to 50\%.
As unimodal models, we use ResNet-18.
The naive late-fusion network classifies the concatenated features
extracted from two ResNet-18 encoders.
Note that in this synthetic setting, HoDEs observe a single image per
sample, whereas late-fusion networks and HeDEs operate on a pair of
different images from the same class.
For all ensembles, we average output logits rather than probabilities
(see Sec.~\ref{sec:appendix_logits_vs_probs}).

Fig.~\ref{fig:motivation_picture} shows how performance changes as the
weaker modality becomes stronger.
The naive late-fusion classifier benefits from two modalities only when
they are nearly balanced, marginally outperforming the stronger
unimodal baseline.
In contrast, deep ensembles outperform the late-fusion baseline across
all imbalance levels.
When the imbalance is moderate, heterogeneous ensembles perform best
by exploiting both modalities.
Under extreme imbalance, homogeneous ensembles of the stronger modality outperform their heterogeneous counterparts, as including the weaker modality degrades the heterogeneous ensemble's performance.

%% file: content/section_4_synthetic_data.tex
\paragraph{Class-aligned setup.} We begin with a unimodal dataset $D = \{(x_1, y_1), \ldots, (x_n, y_n)\}$ consisting of $n$ samples, where $x_i$ are the input data and $y_i$ are the corresponding class labels.
To construct a bimodal dataset, we split $D$ into two non-overlapping subsets $I_1$ and $I_2$ of sizes $s_1$ and $s_2$, respectively, preserving the class distribution; we do not require that $s_1 + s_2 = n$.
Without loss of generality, assume $s_1 \geq s_2$.
A bimodal sample is a triple $(x^1_{i}, x^2_{j}, y)$ where $x^1_i \in I_1$, $x^2_j \in I_2$, and their class labels match.
To form the bimodal dataset, we iterate through $I_1$ and pair each $x^1_i$ with a sample from $I_2$ of the same class $y^1_i$, cycling through the matching-class samples in $I_2$ as needed so that labels always agree. 
The resulting dataset has size $s_1$.
We denote datasets by the fraction of samples available for each modality: $(s_1/n, s_2/n)$.
This construction generalizes straightforwardly to $M > 2$ modalities; we present trimodal results in Section~\ref{subsec:results_synthetic}.
For the test set, each multimodal input consists of $M$ different samples from the same class.
To produce a bimodal dataset with modalities of similar predictive strength, we split $D$ into two equal subsets. 
Keeping one subset fixed while decreasing the size of the other reduces generalization for the weaker modality and makes it converge faster (see Fig.~\ref{fig:motivation_picture}). 
\paragraph{Instance-aligned setup.}
The class-aligned construction is subject to several important limitations. First, varying subset sizes does not allow us to independently control generalization and convergence, but systematically changes both.
One way to overcome this is to inject per-image Gaussian noise into one modality, which slows its convergence and degrades its performance, enabling us to study the regime where the weaker modality converges more slowly (see Appendix~\ref{subsec:appendex_noise_corruption}).
More importantly, the class-aligned construction does not allow us to produce modalities with modality-specific features.
To overcome this, we design a bimodal instance-aligned synthetic multimodal dataset as a set of triples $(G_1(x_i), x_i - G_2(x_i), y_i)$, where $(x_i, y_i) \in D$, and $G_{1,2}$ are Gaussian smoothings with $\sigma_{1,2}$ controlling the frequency bands for each of the modalities.
Feature bands overlap when $\sigma_1 < \sigma_2$, are complementary when $\sigma_1 = \sigma_2$, and have a gap when $\sigma_1 > \sigma_2$.
Both constructions, class- and instance-aligned, can be combined to produce a synthetic multimodal dataset.


%% file: content/section_5_methods.tex
Consider a bimodal dataset $D = \{(x^1_j, x^2_j, y_j)\}_{j=1}^N$, either synthetic or real-world, where $x^m_j$ is a sample from modality $m$ and $y_j$ is the class label. 
For each modality $m = 1, \ldots, M$, we define a $k_m$-th classifier as $g_{k_m}^m = L^m_{k_m} \circ f^m_{k_m}$, where $f^m_{k_m}: \mathbb{R}^{d_i} \to \mathbb{R}^{d_i'}$ is a feature extractor and $L_{k_m}^m: \mathbb{R}^{d_i'} \to \mathbb{R}^C$ is a linear layer projecting features onto $C$-dimensional logit vector. 
To construct a deep ensemble, we train $n_1$ classifiers on modality 1 and $n_2$ on modality 2, each from a different initialization. 
The heterogeneous ensemble averages logits over all $N_{\text{ens}} = n_1 + n_2$ networks:
\begin{gather*}
h_{\text{ens}}(x^1_j, x^2_j) = \frac{1}{N_{\text{ens}}} \left( \sum_{k_1=1}^{n_1} g_{k_1}^{1}(x^1_j) + \sum_{k_2=1}^{n_2} g_{k_2}^{2}(x^2_j) \right).
\end{gather*}
When $n_2 = 0$ (or $n_1 = 0$), this reduces to a homogeneous deep ensemble (HoDE). 
The formulation generalizes straightforwardly to $M > 2$ modalities. 
We find it crucial to average output logits rather than probabilities when training and testing deep ensembles (see Appendix~\ref{sec:appendix_logits_vs_probs} and \citep{tassi_2022_impact, wood_2023_unified}).
Beyond improving accuracy, logit averaging also yields well-calibrated predictions without post-hoc calibration; we present reliability diagrams and Expected Calibration Error analysis in Appendix~\ref{sec:rel_diags_hedes}.


To ensure a fair comparison, we match each ensemble of $N_\text{ens}$ unimodal networks to a late-fusion network with the same modality-specific feature extractors, $\{f^m_{k_m}:m = 1, \ldots, M; k_m = 1, \ldots, n_m\}$, differing only in the classification head. 
The late-fusion network uses a single linear layer mapping the concatenated features $[f^m_{k_m}(x^m_i), m = 1, \ldots, M; k_m = 1, \ldots, n_m]$ to $\mathbb{R}^C$. 
The late-fusion methods we compare against apply various regularization techniques that affect training time but not parameter count. 
A detailed comparison of training cost, memory usage, and parameter counts is provided in Appendix~\ref{sec:appendix_cost_comparisons}.

%% file: content/section_6_results.tex
We compare deep ensembles with existing late-fusion methods on synthetic bimodal datasets generated from CIFAR-100 and ImageNet-1K~\cite{deng_2009_imagenet}. 
We also conduct trimodal experiments on ImageNet-1K. 
As we demonstrate in Sec.~\ref{subsec:results_on_real_data}, our findings on synthetic data generalize to real-world datasets (CREMA-D~\cite{cao_2014_crema}, Sarcasm~\cite{cai_2019_sarcasm_dataset}, Kinetics-400~\cite{kay_2017_kinetics}), and CMU-MOSEI~\cite{zadeh_2018_mosei}.
Throughout, multimodal models (HeDEs and late-fusion networks) use all available modalities and follow the setup from Sec.~\ref{sec:synthetic_data}, while unimodal approaches (single networks and HoDEs) operate on a single modality per model. All baseline results were reproduced using the authors' original code repositories; details are in Appendix~\ref{sec:appendix_training_details}.

%% file: content/section_6_2_number_of_networks.tex
Identifying the optimal heterogeneous deep ensemble of size $N_\text{ens}$ requires searching over all possible allocations of networks to modalities.
A reliable method for estimating the number of models from each modality would be advantageous.

The cross-entropy validation loss $l_i$ of a unimodal classifier trained on modality $X_i$ can be interpreted as the conditional entropy $H(Y|X_i)$, quantifying how much additional information is needed to explain $Y$ given $X_i$.
We propose using the ratio of validation losses to determine the proportion of classifiers for each modality.
Given $M$ input modalities $X_i, i = 1, \ldots, M$, the number of networks $n_i$ trained on modality $X_i$ in a heterogeneous ensemble of size $N_\text{ens}$ is:
\begin{gather}
  n_i = \mathrm{round}\left(N_\text{ens}\frac{1 / l_i}{\sum_j1 / l_j}\right). \label{eq:heuristics}
\end{gather}
To obtain integer counts summing to $N_\text{ens}$, we apply the Largest Remainder Method to $n_i$, breaking ties by smaller $l_i$.
In Appendix~\ref{sec:appendix_additional_results}, we show that our heuristic predicts nearly-optimal configurations across all synthetic and real datasets considered.

%% file: content/section_6_1_results_on_synthetic_data.tex
\begin{figure}[htbp]
    \centering
    \begin{subfigure}[t]{0.45\linewidth}
        \centering
        \includegraphics[width=\linewidth]{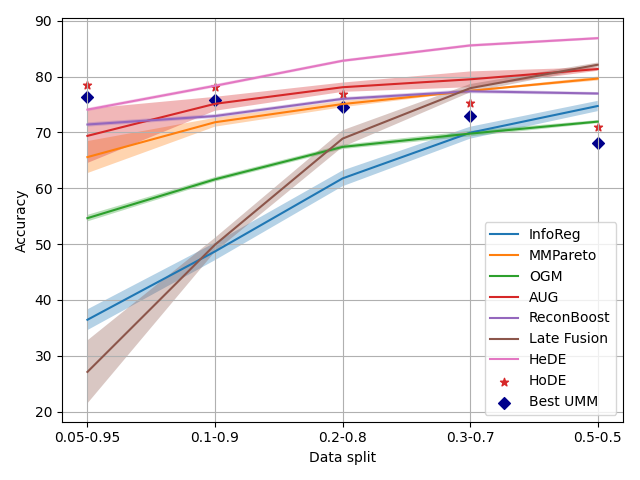}
        \caption{Bimodal CIFAR-100 across data splits}
        \label{figure:mm_cifar_model_size_2}
    \end{subfigure}
    \hfill
    \begin{subfigure}[t]{0.45\linewidth}
        \centering
        \includegraphics[width=\linewidth]{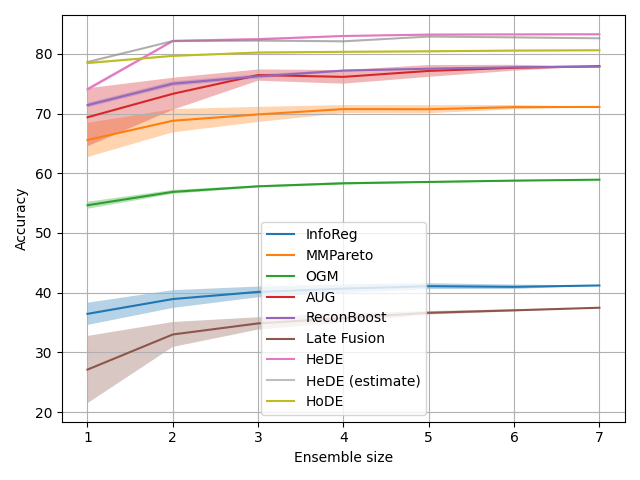}
        \caption{Accuracy vs. ensemble size for the (0.05–0.95) split}
        \label{fig:acc_vs_ens_size_all_methods_0.05_0.95}
    \end{subfigure}
    \hfill
\end{figure}

\paragraph{Bimodal class-aligned setup.}
We compare HeDEs and HoDEs of size 2 with late-fusion baselines specifically designed to address modality imbalance during training, namely AUG \cite{jiang_2025_aug}, InfoReg \cite{huang_2025_inforeg}, MMPareto \cite{wei_2024_mmpareto}, OGM-GE \cite{peng_2022_ogm}, and ReconBoost \cite{hua_2024_reconboost}, on bimodal synthetic CIFAR-100 across five data splits (Fig.~\ref{figure:mm_cifar_model_size_2}).
In the most imbalanced split (0.05–0.95), HoDEs outperform all other approaches.
For more balanced splits, HeDEs yield the best performance. 
The naive late-fusion network without regularization performs on par with the baseline methods and outperforms them for the balanced split. 
Results on bimodal ImageNet-1K are consistent (see Appendix~\ref{appendix_add_results_imagenet}).

Crucially, the advantage of HoDEs under extreme imbalance does not persist as ensemble size grows. 
Fig.~\ref{fig:acc_vs_ens_size_all_methods_0.05_0.95} shows how accuracy changes with ensemble size for the most imbalanced (0.05–0.95) split. 
The horizontal axis indicates the number of late-fusion networks; HeDEs and HoDEs use twice as many unimodal networks for fair parameter comparison. 
Starting from an ensemble size of four, HeDEs outperform HoDEs and all late-fusion methods. 
This pattern is consistent across datasets, including ImageNet-1K (Appendix~\ref{appendix_add_results_imagenet}), CREMA-D (Appendix~\ref{appendix_add_results_crema}), and Kinetics-400 (Section~\ref{subsec:results_on_real_data}): HoDEs lead at small scale under extreme imbalance, while HeDEs take the lead when scaled up.
Our heuristic captures this transition, predicting HoDEs at small sizes and HeDEs at larger ones. 
As shown by the gray \texttt{HeDE (estimate)} line in Figure~\ref{fig:acc_vs_ens_size_all_methods_0.05_0.95}, the heuristic correctly identifies the switch from HoDEs to HeDEs and yields near-optimal allocations at larger ensemble sizes.
Additional results for bimodal CIFAR-100 are provided in Appendix~\ref{appendix_add_results_cifar}.

\paragraph{Bimodal instance-aligned setup.} 
The class-aligned setup allows us to change the convergence speed but cannot separate the features available to the modalities. 
The instance-aligned setup overcomes this shortcoming by assigning low- and high-frequency features to input modalities.
Table~\ref{tab:instance_aligned_cifar_hode} compares the performance of homogeneous, heterogeneous ensembles of size 2, and vanilla late-fusion networks for an instance-aligned dataset with different levels of overlap of frequency bands.
Ensembles outperform the late-fusion baseline for all three configurations of sigmas.
Homogeneous ensembles perform better than heterogeneous counterparts when feature bands do not overlap.
Our heuristic correctly predicts model allocation in the \texttt{Gap} and \texttt{Overlap} configurations, but incorrectly favors the heterogeneous allocation in the \texttt{Exact} case.
Additional results are presented in Appendix~\ref{appendix_add_results_cifar}.
\begin{table}
\centering
\begin{tabular}{rlllll}
\toprule
$\sigma_1/\sigma_2$ & LF & HoDE ($\sigma_1$) & HoDE ($\sigma_2$) & HeDE & Heuristic\\
\midrule
6.0 / 0.4  (Gap) & 75.8 $\pm$ 0.3 & 43.6 $\pm$ 0.2 & \textbf{77.2 $\pm$ 0.2} & 71.9 $\pm$ 0.3 & 77.2 $\pm$ 0.2 \\
2.0 / 2.0  (Exact) & 76.2 $\pm$ 0.2 & 68.3 $\pm$ 0.4 & \textbf{77.4 $\pm$ 0.3} & 76.6 $\pm$ 0.3 & 76.6 $\pm$ 0.3 \\
1.0 / 2.5  (Overlap) & 76.7 $\pm$ 0.4 & 76.2 $\pm$ 0.2 & 77.4 $\pm$ 0.2 & \textbf{78.1 $\pm$ 0.2} & 78.1 $\pm$ 0.2 \\
\bottomrule
\end{tabular}
\caption{Mean accuracy and standard deviation over every distinct choice of members; HoDE(2) is a homogeneous ensemble of two networks of the same band; LF is a single late-fusion network; HeDE are heterogeneous ensembles at $N=2$. The heuristic shows the performance of an ensemble predicted by Formula~\ref{eq:heuristics}.}
\label{tab:instance_aligned_cifar_hode}
\end{table}

\paragraph{Trimodal results.}
\begin{wrapfigure}{r}{0.35\textwidth}
    \centering
    \includegraphics[width=\linewidth]{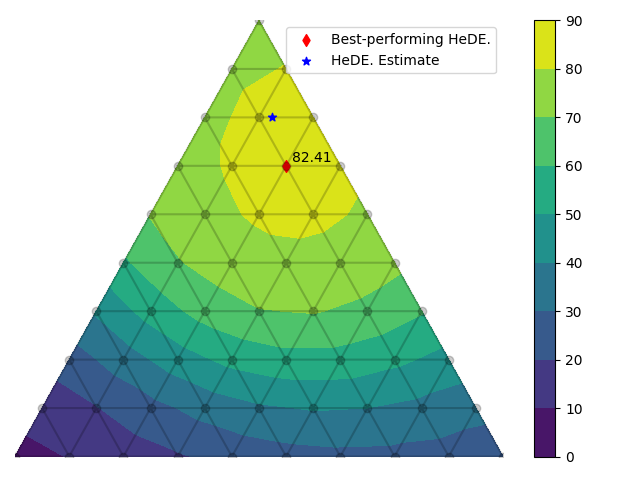}
    \caption{Trimodal ImageNet-1K dataset (0.01–0.03–0.96); ensemble size 9; results on synthetic datasets; red diamond: brute-force optimum; blue star: heuristic estimate}
    \label{fig:mm_in_3_mods}
\end{wrapfigure}
The size of ImageNet-1K allows us to scale the number of modalities while maintaining extreme imbalance between modalities.
Fig.~\ref{fig:mm_in_3_mods} shows the performance of HeDEs of size 9 on a trimodal (0.01–0.03–0.96) split. 
The optimal ensemble lies in the interior of the ternary plot, indicating that HeDEs benefit from including models from every modality. 
Our heuristic closely predicts this configuration. 
As in the bimodal case, HoDEs are preferable for extremely imbalanced splits, while HeDEs achieve the best results when modalities are more balanced (see Appendix~\ref{appendix_add_results_imagenet}).

%% file: content/section_6_3_real_data.tex
In addition to the synthetic datasets considered in the previous section, we conduct extensive experiments on real multimodal datasets.
We selected datasets of varying sizes and different modalities requiring models with different inductive biases.
We use four datasets: CREMA-D (small, audio-visual) \cite{cao_2014_crema}, Sarcasm (small, text-visual) \cite{cai_2019_sarcasm_dataset}, Kinetics-400 (large, audio-visual) \cite{kay_2017_kinetics}, and CMU-MOSEI (medium-size, audio-text-visual).
\paragraph{CREMA-D.}The CREMA-D dataset is widely used to assess late-fusion networks under modality imbalance \cite{fan_2023_pmr, hua_2024_reconboost, huang_2025_inforeg, jiang_2025_aug, kontras_2025_mcr, li_2023_agm, peng_2022_ogm, yang_2024_lfm}. However, multiple existing methods report results under data leakage, leading to inflated performance (see Appendix~\ref{appendix_add_results_crema}). In Table~\ref{tab:clean_results_crema}, we report accuracies on our clean split without data leakage. Fig.~\ref{fig:crema_ens_size_2_heuristics} shows that for ensembles of size 2, the HeDE outperforms both audio and video HoDEs, and our heuristic closely predicts the optimal modality allocation.
\begin{figure}[htbp]
\begin{minipage}[t]{0.48\linewidth}
\vspace{0pt}
\centering
\includegraphics[width=\linewidth]{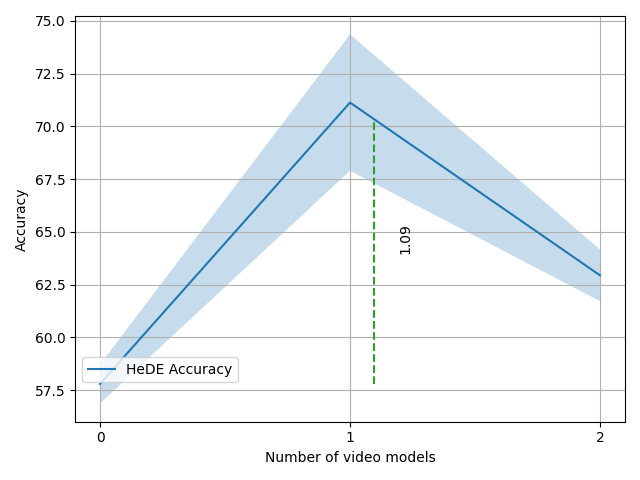}
\caption{Mean accuracy with standard deviation across 10 ensembles of size 2. The dashed line indicates the heuristic prediction.}
\label{fig:crema_ens_size_2_heuristics}
\end{minipage}%
\hfill
\begin{minipage}[t]{0.4\linewidth}
\vspace{15pt}
\centering
\begin{tabular}{r|l}
\toprule
Method & Accuracy \\
\hline
Vanilla LF & 56.7 $\pm$ 1.6 \\
OGM & 56.7 $\pm$ 1.4 \\
LFM & 63.2 $\pm$ 1.7 \\
AUG & 65.0 $\pm$ 1.6 \\
InfoReg & 59.7 $\pm$ 1.1 \\
MMPareto & 61.2 $\pm$ 0.8 \\
ReconBoost & 61.7 $\pm$ 0.7\\
Audio HoDE & 58.2 $\pm$ 1.4 \\
Video HoDE & 62.8 $\pm$ 1.5\\
HeDE & \textbf{71.1 $\pm$ 3.1} \\
\bottomrule
\end{tabular}
\captionof{table}{Mean accuracy and standard deviation across 5 runs on the clean CREMA-D split}
\label{tab:clean_results_crema}
\end{minipage}
\end{figure}

As in the synthetic setting, we can control modality strength by varying the amount of training data. 
Results are consistent: HoDEs are preferable under heavy imbalance, HeDEs in all other cases, and our heuristic correctly predicts the transition. 
Detailed results are in Appendix~\ref{appendix_add_results_crema}.

\paragraph{Sarcasm.}We compare deep ensembles to late-fusion baselines on the Sarcasm dataset~\cite{cai_2019_sarcasm_dataset}. 
For all approaches, we use BERT-Base~\cite{devlin_2019_bert} as the text encoder and ResNet-50 as the image encoder, both fine-tuned from pre-trained checkpoints. 
As shown in Fig.~\ref{fig:sarcasm_fine_tune_pre_trained}, HeDEs perform on par with the best late-fusion methods at small ensemble sizes and outperform all late-fusion approaches starting from ensemble size 6 (equivalent to three late-fusion networks).
At larger ensemble sizes, HeDEs maintain their advantage, though accuracy eventually saturates for all methods. 
HeDEs also consistently outperform HoDEs, even at ensemble size 2. 
Our heuristic reliably predicts the optimal ensemble configuration on this dataset as well. 
Detailed results, including alternative training strategies, are in Appendix~\ref{appendix_add_results_sarcasm}.

\begin{figure}[htbp]
\centering
\begin{subfigure}[t]{0.38\linewidth}
\centering
\includegraphics[width=\linewidth]{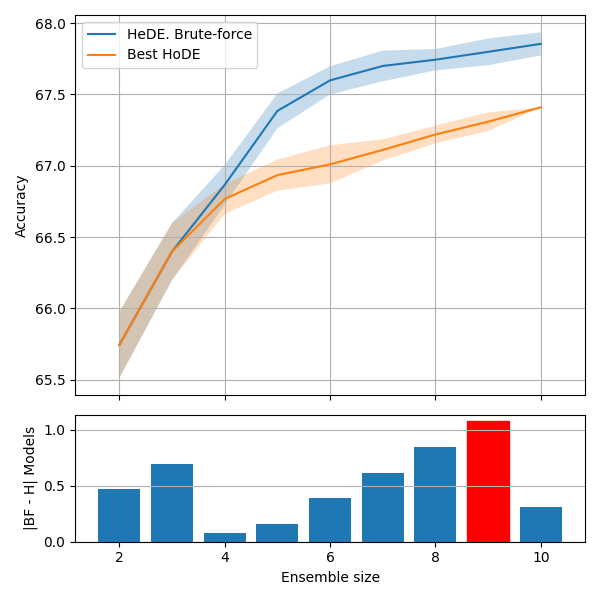}
\caption{Top panel shows accuracy of HeDEs and HoDEs across ensemble sizes. Bottom panel: discrepancy between brute-force and heuristic allocations}
\label{fig:k400_hede_vs_hode}
\end{subfigure}
\hfill
\begin{subfigure}[t]{0.48\linewidth}
\centering
\includegraphics[width=\linewidth]{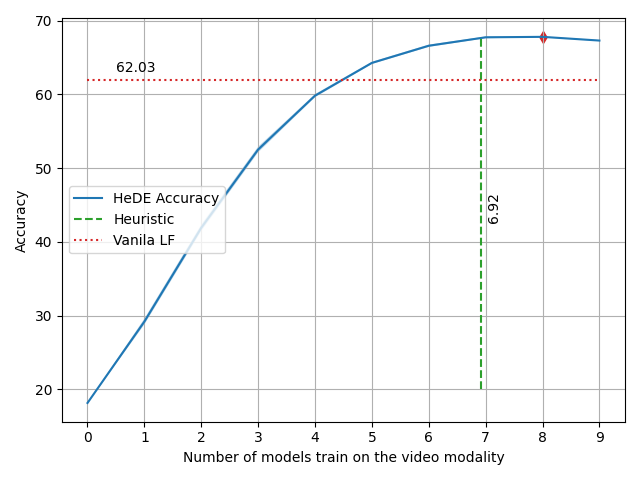}
\caption{HeDE accuracy vs. number of video networks for ensemble size 9; red diamond: brute-force; dashed line: heuristic}
\label{fig:k400_heuristics_ens_size_9}
\end{subfigure}
\end{figure}

\paragraph{Kinetics-400.}Kinetics-400~\cite{kay_2017_kinetics} is a large-scale video classification dataset where the audio modality is considerably weaker than the visual modality~\cite{du_2023_umt_and_ume, wang_2020_what_make_training_hard}. 
Fig.~\ref{fig:k400_hede_vs_hode} (top panel) shows accuracy across ensemble sizes: starting from size 4, HeDEs consistently outperform HoDEs, mirroring the scaling pattern observed on synthetic datasets (Sec.~\ref{subsec:results_synthetic}). 
The bottom panel shows that the difference between the brute-force-optimal and heuristic-predicted number of video networks remains small across all ensemble sizes. 
Fig.~\ref{fig:k400_heuristics_ens_size_9} details the largest such difference (ensemble size 9, corresponding to the red bar in Fig.~\ref{fig:k400_hede_vs_hode}, bottom panel): the heuristic's prediction is near-optimal, confirming that it produces reliable allocation estimates for heavily imbalanced real-world datasets. 
For comparison, Fig.~\ref{fig:k400_heuristics_ens_size_9} includes the performance of the vanilla late-fusion baseline (red dotted line). 
Training details are provided in Appendix~\ref{sec:appendix_training_details}.

\paragraph{CMU-MOSEI.} 
We tested heterogeneous deep ensembles on the trimodal CMU-MOSEI dataset~\cite{zadeh_2018_mosei}.
Following the standard protocol for this benchmark, we fine-tune per-modality GRU encoders on the provided pre-extracted features \cite{cho_2014_rnns}.
We convert the original sentiment scores into binary and 7-class classification tasks, referred to as MOSEI-2 and MOSEI-7, respectively, by taking the sign of the score (excluding zero) and by rounding it.
Table~\ref{tab:mosei} shows that deep ensembles fall short of the naive late-fusion baseline for the binary classification task.
Moreover, our heuristic favors text and video encoders, whereas brute-force allocation skews toward a single textual modality.
Arguably, our heuristic cannot reliably predict the model's allocation when the ensembles lack diversity and the optimal solution favors a dominant modality.
It is, however, notable that the vanilla late-fusion network is still unable to outperform the best possible ensemble.
\begin{table}[t]
\centering
\begin{tabular}{llcccccc}
\toprule
Task & $N$ & Heuristic & HeDE & LF ens. & BF alloc. & BF acc. \\
\midrule
MOSEI-2 & 3 & $[1,1,1]$ & $77.2 \pm 0.5$ & \textbf{80.0 $\pm$ 0.4} & $[3,0,0]$ & $80.7 \pm 0.3$ \\
MOSEI-2 & 6 & $[2,1,3]$ & $79.5 \pm 0.3$ & \textbf{80.8 $\pm$ 0.3} & $[5,0,1]$ & $81.0 \pm 0.2$ \\
MOSEI-7 & 3 & $[1,0,2]$ & \textbf{47.2 $\pm$ 0.5} & $46.0 \pm 0.5$ & $[3,0,0]$ & $47.9 \pm 0.4$ \\
MOSEI-7 & 6 & $[2,1,3]$ & \textbf{48.4 $\pm$ 0.4} & $47.7 \pm 0.3$ & $[5,0,1]$ & $48.6 \pm 0.3$ \\
\bottomrule
\end{tabular}
\caption{Results on CMU-MOSEI; allocations show the number of [\texttt{Text, Audio, Video}] models within an ensemble.}
\label{tab:mosei}
\end{table}

%% file: content/section_6_4_DE_vs_interfusion_vs_I2M2.tex
In this section we compare deep ensembles to intermediate-fusion networks and hybrid approaches that combine unimodal and multimodal models. To construct an intermediate-fusion network, we take a late-fusion network (Sec.~\ref{sec:deep_ens_and_late_fusion}) and incorporate MMTM fusion blocks \cite{Joze_2020_MMTM}, progressively inserting them at earlier layers of the network (MMTM-1 after layer 4; MMTM-2 after layers 3 and 4; see Table~\ref{table:mmtm}). 
All intermediate-fusion networks are trained from scratch.
We also train ViTs \cite{dosovitskiy_2021_vit} on ImageNet with cross-attention \cite{tan_2019_lxmert, tsai_2019_MULT} as fusion blocks; see Appendix~\ref{sec:appendix_interfusion_and_i2m2} for results.
Table~\ref{table:mmtm} shows that adding one or two intermediate fusion blocks yields marginal improvement over the vanilla late-fusion baseline for CREMA-D. In all cases, deep ensembles outperform intermediate-fusion networks.
\begin{table}[htbp]
\centering
\begin{tabular}{l|l|l|l|l|l}
\toprule
    Dataset & LF & MMTM-1 & MMTM-2 & HoDE(2) & HeDE(2) \\
    \hline
    C-100 (0.05 -- 0.95) & 34.0 $\pm$ 1.5 & 35.3 $\pm$ 0.6 & 33.6 $\pm$ 0.3 & \textbf{78.5 $\pm$ 0.2} & 74.1 $\pm$ 0.3 \\
    C-100 (0.5 -- 0.5) & 82.5 $\pm$ 0.4 & 82.4 $\pm$ 0.2 & 82.1 $\pm$ 0.3 & 71.6 $\pm$ 0.2 & \textbf{86.9 $\pm$ 0.2}\\
    CREMA-D & 56.7 $\pm$ 1.6 & 58.7 $\pm$ 0.9 & 58.9 $\pm$ 1.3  & 62.8 $\pm$ 1.5 & \textbf{71.2 $\pm$ 3.1} \\
\bottomrule
\end{tabular}
\caption{Accuracy of intermediate-fusion classifiers on synthetic and real-world datasets; \texttt{MMTM-X}: network with $X$ MMTM blocks inserted at progressively earlier layers; \texttt{HoDE(2)}: homogeneous ensemble of size two trained on the stronger modality}
\label{table:mmtm}
\end{table}

We also compare deep ensembles to I2M2 \cite{madaan_2024_i2m2}, a hybrid approach that averages the log-probabilities of a multimodal network with those of unimodal networks. 
Note that log-probability averaging yields the same classification accuracy as logit averaging. 
We use a vanilla late-fusion network as the multimodal component, ensuring equal parameter count with HeDE(4). 
The results from Table~\ref{table:de_vs_i2m2} show that deep ensembles outperform I2M2. 
In Appendix~\ref{sec:appendix_interfusion_and_i2m2}, we provide additional results using both late-fusion and early-fusion networks as the multimodal component.
\begin{table}[htbp]
\centering
\begin{tabular}{l|l|l}
\toprule
    Dataset & HeDE(4) & I2M2 \\
    \hline
    CIFAR-100 (0.05 -- 0.95) & \textbf{82.0 $\pm$ 0.2} & 74.9 $\pm$ 0.5\\
    CIFAR-100 (0.5 -- 0.5) & \textbf{88.9 $\pm$ 0.2} & 87.9 $\pm$ 0.2 \\
    ImageNet-1K (0.01 -- 0.99) & \textbf{77.7 $\pm$ 0.1} & 62.4 $\pm$ 0.2 \\
    ImageNet-1K (0.5 -- 0.5) & \textbf{88.5 $\pm$ 0.1} & 88.0 $\pm$ 0.1 \\
    CREMA-D & \textbf{75.3 $\pm$ 0.9} & 70.1 $\pm$ 2.0 \\
\bottomrule
\end{tabular}
\caption{Deep ensembles outperform hybrid approaches on synthetic and real-world datasets at equal parameter count.}
\label{table:de_vs_i2m2}
\end{table}

The choice of a vanilla late-fusion network as the multimodal component of I2M2 might be suboptimal. 
However, as we show in Appendix~\ref{sec:appendix_interfusion_and_i2m2}, adopting early fusion instead does not help I2M2 to outperform deep ensembles. 
Similarly, alternative fusion techniques could improve over the MMTM baseline. 
Notably, any such alternative would need to perform substantially better to close the gap between fusion approaches and deep ensembles. 
While other fusion techniques or multimodal architectures may narrow this gap, the consistent advantage of deep ensembles across all methods and datasets we evaluated suggests that this advantage is robust. 

%% file: content/section_7_scaling_laws.tex
In this section, we fit scaling laws to estimate how the test loss of heterogeneous deep ensembles changes as a function of ensemble size. 
Our goals are twofold: first, to show that for \textit{any} fixed proportion of unimodal models in HeDEs, a power law can be fitted to the test loss as a function of ensemble size; second, to show that the asymptotic test loss of the ensemble can be reliably predicted from unimodal model losses. 
We present additional results in Appendix~\ref{sec:appendix_scaling_laws}.

\begin{figure}[htbp]
\centering
\begin{subfigure}[t]{0.48\linewidth}
    \centering
    \includegraphics[width=1.\linewidth]{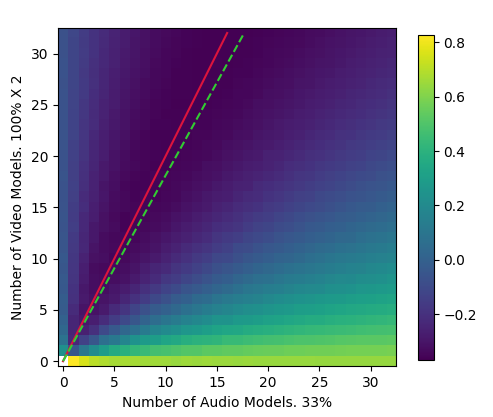}
    \caption{Test log-loss for HeDEs on CREMA-D; red line: optimal proportion; green dashed: heuristic prediction}
    \label{fig:cream_log_loss_vs_ens_size}
\end{subfigure}
\hfill
\begin{subfigure}[t]{0.4\linewidth}
    \centering
    \includegraphics[width=\linewidth]{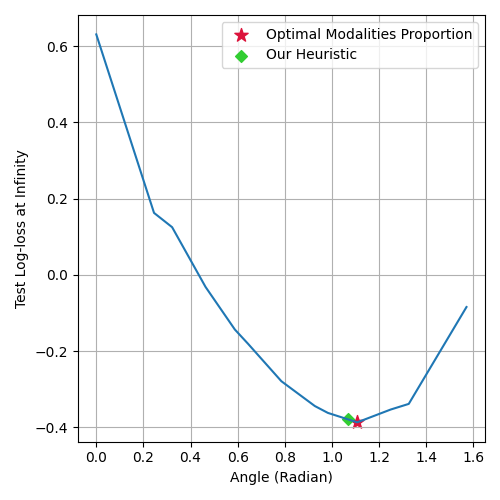}
    \caption{Asymptotic log-loss $L_{\infty}(\theta)$ vs.\ angle $\theta$; red star: optimal; green diamond: heuristic}
    \label{fig:loss_vs_angle}
\end{subfigure}
\end{figure}

To emulate modalities with different predictive strength on CREMA-D, we train audio and video unimodal networks using 10\%, 33\%, 50\%, and 100\% of the original dataset. Additionally, to add stronger visual modalities, we train visual unimodal networks on the entire dataset but use either 2 or 4 video frames as input. In total, this gives us 4 audio and 6 visual modalities, resulting in 24 bimodal scenarios. In Fig.~\ref{fig:cream_log_loss_vs_ens_size}, we show the logarithmic test loss of heterogeneous ensembles for one representative scenario: audio networks trained on 33\% of the data, video networks trained on the full dataset with 2 input frames.


We can fix the proportion of unimodal models and scale the ensemble size accordingly. 
We parameterize this proportion by an angle $\theta\in\left[0,\frac{\pi}{2}\right]$.
In Fig.~\ref{fig:cream_log_loss_vs_ens_size}, each ray from the origin corresponds to a fixed proportion, and all ensembles along that ray share it. 
For instance, horizontal and vertical rays correspond to audio and video HoDEs, respectively, while a ray at $\pi/4$ represents HeDEs with an equal number of audio and video models.

Scaling up the number of models at a fixed proportion $\theta$ allows us to fit power laws of the form $L_N(\theta) = L_{\infty}(\theta) + \frac{A(\theta)}{N^{\alpha(\theta)}}$, where $N$ is the ensemble size and $L_{\infty}(\theta)$ is the asymptotic test loss. We find that power laws of this form fit well across all 24 modality combinations and angles; we refer to Appendix~\ref{sec:appendix_scaling_laws} for examples. Furthermore, our heuristic from Sec.~\ref{subsec:number_of_networks} reliably predicts the optimal proportion, $\theta_{\text{opt}} = \arg\min_{\theta}L_{\infty}(\theta)$. We demonstrate this in Fig.~\ref{fig:cream_log_loss_vs_ens_size}, showing $\theta_{\text{opt}}$ with a solid red line and the heuristic prediction with a dashed green line. Fig.~\ref{fig:loss_vs_angle} shows how the asymptotic log-loss $L_{\infty}(\theta)$ depends on the angle for the same pair of modalities.



%% file: content/section_8_discussion.tex
%
%

Although deep ensembles do not explicitly fuse unimodal features, it remains an open question whether fusion networks can be designed to match their performance. Our comparison with intermediate-fusion (MMTM, cross-attention) and hybrid (I2M2) approaches is a first step, but a broader evaluation across fusion techniques and multimodal architectures, including large-scale multimodal models, is needed. 
While the heuristic we propose in Sec.~\ref{subsec:number_of_networks} performs reliably across our experiments, its theoretical basis remains unclear. 
Addressing these gaps could guide the design of improved ensemble selection and aggregation methods.

Typically, overparametrized models assign high probabilities to correctly predicted samples while keeping considerably smaller probabilities for incorrectly classified instances.
Unimodal models tend to confuse a given sample with different classes, depending on the modality they were trained on.
When some modalities make incorrect predictions, larger-magnitude logits from other modalities with correct predictions can dominate the final ensemble prediction.
Heterogeneous deep ensembles thus benefit from multiple modalities due to the different errors made by individual encoders.

Although our instance-aligned setup from Section \ref{sec:synthetic_data} can produce modalities with unique features, some real-world benchmarks such as Hateful Memes \cite{kiela_2020_hateful_memes} require both modalities to be processed simultaneously to make a correct binary prediction. 
In our experiments, unimodal image and text models yield 51.4\% and 71.3\% accuracy, respectively. 
The late-fusion network achieves 72.4\% accuracy, while the image and text encoders validated individually achieve 53.8\% and 72.2\% accuracy, respectively. 
Although the late-fusion baseline outperforms deep ensembles on this benchmark, it remains unclear whether the late-fusion network learns from both modalities or whether the image encoder regularizes the text modality, allowing a longer search for a better solution.

Validating heterogeneous ensembles on other benchmarks and modality sets, such as RGB images, depth estimates, angles, and edges, is a promising direction for better understanding the applicability of heterogeneous ensembles \cite{zamir_2018_taskonomy}. 
Deep ensembles can be easily adapted to scenarios with missing input modalities by simply avoiding inference on those modalities.
Comparing deep ensembles under missing modalities with other approaches was, however, out of the scope of the present work.

As a methodological study on multimodal classification, this work does not introduce direct societal risks; however, improved multimodal classifiers could be applied in sensitive domains such as surveillance or medical diagnosis, where careful deployment considerations apply.

%% file: content/section_9_conclusion.tex
This work studies deep ensembles for multimodal data classification. 
Our approach trains unimodal networks independently and selects the number of networks per modality according to the heuristic proposed in Sec.~\ref{subsec:number_of_networks}. 
We demonstrate that heterogeneous deep ensembles outperform strong late-fusion baselines, intermediate-fusion, and hybrid approaches on a variety of synthetic and real datasets. 
Under extreme modality imbalance, homogeneous ensembles are preferable at small ensemble sizes; as the ensemble grows, heterogeneous ensembles effectively leverage logits from weaker modalities and take the lead — a transition that our heuristic reliably predicts.
In all other cases, heterogeneous deep ensembles outperform their homogeneous counterparts. 
We also introduce a synthetic multimodal dataset construction that provides control over the predictive strength of individual modalities; our findings are consistent across synthetic and real-world data, suggesting that the synthetic setting captures relevant properties of real-world multimodal learning. 
Finally, we establish scaling laws on both synthetic and real datasets that predict asymptotic test loss from unimodal model performance, with regression quality remaining stable even when using single unimodal models for prediction.

%% file: content/section_010_acknowledgments.tex
I.B. thanks Maria Ponomareva for her support, and Alexander Binder, Lalith Manjunath, Callum Murphy-Barltrop, Andrei Politov, and Martin Weigert for valuable discussions.
I.B. acknowledges the financial support by the Federal Ministry of Research, Technology and Space of Germany and by S\"{a}chsische Staatsministerium f\"{u}r Wissenschaft, Kultur und Tourismus in the programme Center of Excellence for AI-research „Center for Scalable Data Analytics and Artificial Intelligence Dresden/Leipzig“, project identification number: ScaDS.AI. The authors gratefully acknowledge the computing time made available to them on the high-performance computer at the NHR Center of TU Dresden. This center is jointly supported by the Federal Ministry of Research, Technology and Space of Germany and the state governments participating in the NHR (www.nhr-verein.de/unsere-partner).

%% file: content/section_appendix_synthetic_dataset.tex
Our class-aligned construction pairs same-class samples from a single unimodal source, so each modality independently carries features sufficient to predict the class label.
This is the key property shared with real-life instance-aligned multimodal datasets: in both settings, unimodal classifiers are well-defined and can exhibit different generalization and convergence rates, which is the precondition for studying modality imbalance.

The construction differs from real-life datasets in one important respect. 
In instance-aligned data, each modality may possess unique information about the object that the other modality lacks \cite{baltrusaitis_2019_mml_survey, liang_2024_foundations_survey_on_mm_ml}. 
In our construction, because both modality inputs are independently drawn from the same unimodal source conditioned on the class label, the information they carry about the label almost entirely overlaps. 

To overcome this shortcoming, we either use the instance-aligned setup as described in Section \ref{sec:synthetic_data} or add fixed per-sample Gaussian noise. The former allows to separate the feature sets used by the synthetic modalities, while the latter helps to control the convergence speed and reduce the strength of the corrupted modality.

In this work, our goal is to compare deep ensembles against imbalance-aware late-fusion methods under modality imbalance and to evaluate a heuristic for allocating networks across modalities in an ensemble. 
These questions depend on per-modality generalization and convergence characteristics, which our construction preserves and allows us to vary in a controlled manner. 
Our findings hold consistently across synthetic class-aligned (Sections~\ref{subsec:results_synthetic}), real-world instance-aligned (Section~\ref{subsec:results_on_real_data}), and real-world datasets with controlled modality imbalance (Appendix \ref{subsec:appendxi_heuristic}).

%% file: content/section_appendix_a_additional_results.tex
\subsection{CIFAR-100}
\label{appendix_add_results_cifar}
\paragraph{Class-aligned setup.}
In this section we provide additional results for the bimodal synthetic CIFAR-100 dataset.
Table~\ref{table:cifar_2_mods_lf_vs_hodes_vs_hede} compares accuracies for the naive late-fusion network, homogeneous ensembles of the weaker and stronger modalities, and heterogeneous ensembles across modality imbalance levels we considered.
\begin{table*}[htbp]
    \centering
    \begin{tabular}{r|l|l|l|l}
    \toprule
        Data Split&Late Fusion&HoDE 1&HoDE 2&HeDE \\
        \hline
        0.05 -- 0.95 & 34.0 $\pm$ 1.5 & 24.3 $\pm$ 0.4 & \textbf{78.5 $\pm$ 0.2} & 74.1 $\pm$ 0.3 \\
        0.1 -- 0.9   & 51.5 $\pm$ 1.6 & 39.2 $\pm$ 0.3 & 77.9 $\pm$ 0.2 & \textbf{78.4 $\pm$ 0.3} \\
        0.2 -- 0.8   & 68.1 $\pm$ 1.4 & 55.1 $\pm$ 0.3 & 76.8 $\pm$ 0.3 & \textbf{82.8 $\pm$ 0.3} \\
        0.3 -- 0.7   & 78.6 $\pm$ 0.4 & 62.9 $\pm$ 0.4 & 75.2 $\pm$ 0.2 & \textbf{85.6 $\pm$ 0.3} \\
        0.5 -- 0.5   & 82.5 $\pm$ 0.4 & 71.5 $\pm$ 0.2 & 70.5 $\pm$ 0.3 & \textbf{86.9 $\pm$ 0.2} \\
    \bottomrule
    \end{tabular}
    \caption{
    Accuracy on the bimodal synthetic CIFAR-100 dataset across modality imbalance levels, comparing the naive late-fusion network, homogeneous deep ensembles of the weaker (\texttt{HoDE 1}) and stronger (\texttt{HoDE 2}) modality, and the heterogeneous deep ensemble (\texttt{HeDE}) allocated by our heuristic.
    All ensembles have size $N_\text{ens} = 2$.
    }
    \label{table:cifar_2_mods_lf_vs_hodes_vs_hede}
\end{table*}

To understand why late-fusion networks underperform, we analyze the individual encoders within the trained late-fusion network.
We use features extracted by an encoder and project them using the corresponding half of the trained weights in the linear layer.
In Table~\ref{table:cifar_2_mods_encoders_vs_resnets} we compare the performance of the individual encoders with that of unimodal networks trained on either weaker or stronger modality.
Interestingly, under extreme modality imbalance (0.05 -- 0.95) both encoders show similar results. When the modality imbalance is slightly relaxed, the encoder trained on the weaker modality outperforms its counterpart trained on the stronger modality. 
Under milder imbalance, encoders show expected results: an encoder trained on the stronger modality performs better.
\begin{table*}[htbp]
    \centering
    \begin{tabular}{r|l|l|l|l}
    \toprule
        Data Split &LF Enc. WM& LF Enc. SM & RN Mod. WM & RN Mod. SM \\
        \hline
        0.05 – 0.95 & 18.4 $\pm$ 4.4 & 18.9 $\pm$ 5.4 & 22.9 $\pm$ 0.4 & \textbf{76.3 $\pm$ 0.3} \\
        0.1 – 0.9   & 39.8 $\pm$ 0.5 & 37.6 $\pm$ 1.2 & 36.8 $\pm$ 0.4 & \textbf{75.8 $\pm$ 0.3} \\
        0.2 – 0.8   & 50.8 $\pm$ 0.5 & 55.9 $\pm$ 1.2 & 51.8 $\pm$ 0.5 & \textbf{74.5 $\pm$ 0.3} \\
        0.3 – 0.7   & 57.0 $\pm$ 0.5 & 63.3 $\pm$ 0.5 & 59.9 $\pm$ 0.4 & \textbf{73.0 $\pm$ 0.3} \\
        0.5 – 0.5   & 60.8 $\pm$ 0.9 & 61.8 $\pm$ 0.5 & \textbf{68.6 $\pm$ 0.3} & 68.1 $\pm$ 0.4 \\
    \bottomrule
    \end{tabular}
    \caption{Top-1 accuracy (\%) of the two encoders within a trained bimodal late-fusion network, evaluated by projecting their features through the corresponding half of the fusion layer, compared against unimodal ResNets trained separately on each modality. All models trained on the synthetic bimodal CIFAR-100 (see Section~\ref{sec:synthetic_data}). \texttt{LF Enc.} denotes a late-fusion encoder, \texttt{RN Mod.} a unimodal ResNet; \texttt{WM} and \texttt{SM} the weaker and stronger modality.
    }
    \label{table:cifar_2_mods_encoders_vs_resnets}
\end{table*}

\begin{figure}[htbp]
\centering
\begin{subfigure}[t]{0.48\linewidth}
    \centering
    \includegraphics[width=1.\linewidth]{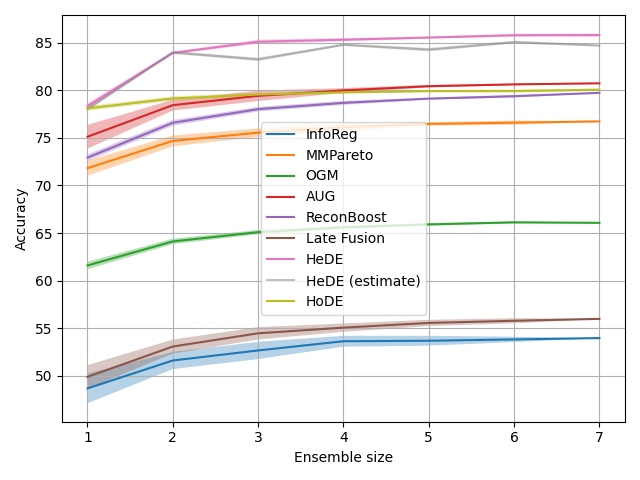}
    \caption{CIFAR-100 (0.1 -- 0.9).}
    \label{fig:acc_vs_ens_size_all_methods_0.1_0.9}
\end{subfigure}
\hfill
\begin{subfigure}[t]{0.48\linewidth}
    \centering
    \includegraphics[width=\linewidth]{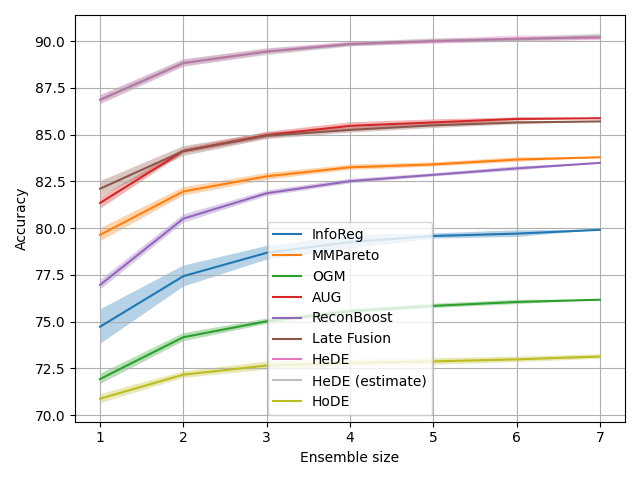}
    \caption{CIFAR-100 (0.5 -- 0.5).}
    \label{fig:acc_vs_ens_size_all_methods_0.5_0.5}
\end{subfigure}
\caption{Under milder modality imbalance, HeDEs significantly outperform state-of-the-art late-fusion methods designed to handle modality imbalance.}
\label{fig:acc_vs_ens_size_cifar_two_example}
\end{figure}

As shown in Section~\ref{subsec:results_synthetic}, under extreme modality imbalance, heterogeneous deep ensembles outperform their homogeneous counterparts when the ensemble size is scaled up.
In Figure~\ref{fig:acc_vs_ens_size_cifar_two_example}, we show that HeDEs significantly outperform other methods specifically designed to address the modality imbalance starting from ensemble size 2.
Notably, our heuristic predicts close-to-optimal choice of the number of unimodal models in the heterogeneous deep ensembles (gray curves on Figs.~\ref{fig:acc_vs_ens_size_all_methods_0.1_0.9} and \ref{fig:acc_vs_ens_size_all_methods_0.5_0.5}).

\begin{figure}
    \centering
    \includegraphics[width=0.95\linewidth]{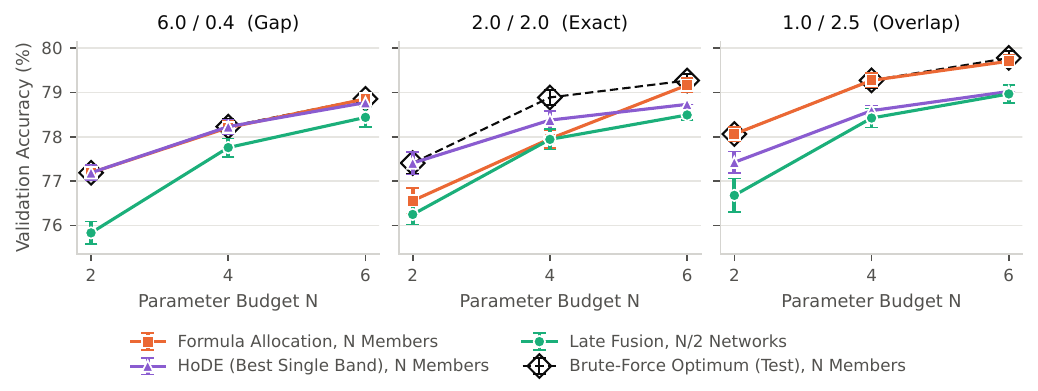}
    \caption{Accuracies for homogeneous, heterogeneous deep ensembles together with the ensembles of late-fusion networks, and brute-force optimal solution}
    \label{fig:instance_aligned_acc_vs_N}
\end{figure}

\paragraph{Instance-aligned setup.}
For the instance-aligned modalities with different band overlaps, we compare the performance of heterogeneous deep ensembles to that of homogeneous counterparts and vanilla late-fusion baselines.
Figure~\ref{fig:instance_aligned_acc_vs_N} compares the performance of deep ensembles, homogeneous and heterogeneous ones, to that of ensembles of naive late-fusion networks.
Our results from the class-aligned setup generalize to the instance-aligned construction.
When the feature bands have a gap, $\sigma_1 = 6.0, \sigma_2 = 0.4$, our heuristic predict a transition from homogeneous ensembles ($N = 2, 4$) to a heterogeneous ensemble ($N = 6$).
For the complementary feature bands (\texttt{Exact} in Figure~\ref{fig:instance_aligned_acc_vs_N}), heterogeneous ensembles outperform other competitors and approach the brute-force optimal solution when the ensemble size scales up.
In the \texttt{Overlap} case, our heuristic predicts close-to-optimal allocations of models; see also Table~\ref{tab:formula_vs_bruteforce}.
\begin{table}
\centering
\begin{tabular}{lcccrrl}
\toprule
$\sigma$-set & $N$ & Heuristic & Brute-force & Heuristic acc. & Best acc. & Match \\
\midrule
Gap     & 2 & $[0,2]$ & $[0,2]$ & 77.19 & 77.19 & yes \\
Gap     & 4 & $[1,3]$ & $[0,4]$ & 78.21 & 78.23 & no ($\Delta = -0.02$) \\
Gap     & 6 & $[1,5]$ & $[1,5]$ & 78.85 & 78.85 & yes \\
\midrule
Exact   & 2 & $[1,1]$ & $[0,2]$ & 76.56 & 77.41 & no ($\Delta = -0.85$) \\
Exact   & 4 & $[2,2]$ & $[1,3]$ & 77.95 & 78.89 & no ($\Delta = -0.94$) \\
Exact   & 6 & $[2,4]$ & $[1,5]$ & 79.16 & 79.27 & no ($\Delta = -0.11$) \\
\midrule
Overlap & 2 & $[1,1]$ & $[1,1]$ & 78.06 & 78.06 & yes \\
Overlap & 4 & $[2,2]$ & $[2,2]$ & 79.27 & 79.27 & yes \\
Overlap & 6 & $[3,3]$ & $[2,4]$ & 79.70 & 79.78 & no ($\Delta = -0.08$) \\
\bottomrule
\end{tabular}
\caption{Instance-aligned-CIFAR-100 validation accuracy (\%) of the formula allocation compared with the brute-force best allocation at the same budget $N$. Allocations are $[n_{\sigma_1}, n_{\sigma_2}]$, the number of networks per band. The brute-force best is chosen by enumerating every allocation on the validation set. Accuracies are means over every distinct choice of members; the difference in the last column is formula minus best.}
\label{tab:formula_vs_bruteforce}
\end{table}


\subsection{ImageNet-1K}
\label{appendix_add_results_imagenet}
\paragraph{Bimodal setup.}Results on bimodal ImageNet-1K are consistent with CIFAR-100.
Table~\ref{table:in_2_mods_lf_vs_hodes_vs_hede} shows that when modalities are imbalanced, (0.01 -- 0.99) and (0.03 -- 0.97) modality splits, homogeneous ensembles outperform heterogeneous counterparts.
When the imbalance is less pronounced, heterogeneous ensembles outperform other methods.
\begin{table*}[htbp]
    \centering
    \begin{tabular}{r|l|l|l|l}
    \toprule
        Data Split&Late Fusion&HoDE 1&HoDE 2&HeDE \\
        \hline
        0.01 – 0.99  & 10.8 $\pm$ 0.1 & 6.9 $\pm$ 0.1  & \textbf{76.7 $\pm$ 0.1} & 61.7 $\pm$ 0.2 \\
        0.03 – 0.97  & 31.4 $\pm$ 0.3 & 22.7 $\pm$ 0.2 & \textbf{76.5 $\pm$ 0.1} & 72.7 $\pm$ 0.2 \\
        0.1 – 0.9    & 62.6 $\pm$ 0.3 & 49.6 $\pm$ 0.2 & 76.2 $\pm$ 0.1 & \textbf{82.3 $\pm$ 0.1} \\
        0.33 – 0.67  & 83.6 $\pm$ 0.1 & 67.7 $\pm$ 0.1 & 74.0 $\pm$ 0.1 & \textbf{86.4 $\pm$ 0.1} \\
        0.5 – 0.5    & 84.3 $\pm$ 0.1 & 71.7 $\pm$ 0.1 & 71.8 $\pm$ 0.1 & \textbf{86.8 $\pm$ 0.1} \\
    \bottomrule    
    \end{tabular}
    \caption{Performance of the naive late-fusion classifier, homogeneous deep ensembles for both modalities, and heterogeneous deep ensembles.
    All models trained on the synthetic version of the bimodal ImageNet-1K dataset.}
    \label{table:in_2_mods_lf_vs_hodes_vs_hede}
\end{table*}

In Table~\ref{table:in_2_mods_encoders_vs_resnets} we compare the performance of individual encoders of pre-trained vanilla neural networks with that of unimodal models trained on weaker and stronger modalities, following the setup from the previous section (Sec.~\ref{appendix_add_results_cifar}, also Table~\ref{table:cifar_2_mods_encoders_vs_resnets}).
As in the case of the bimodal CIFAR-100 dataset, encoders within the late-fusion network exhibit different performance correlated with modality strength only starting from a certain modality split, (0.1 -- 0.9) in the case of the bimodal ImageNet dataset. 
Interestingly, in the most extreme modality imbalance we tested, the encoder exposed to the weaker modality achieves higher accuracy.
Furthermore, accuracy achieved by an independent encoder within the late-fusion network trained on the weaker modality is higher than that of the unimodal network trained on the same modality (9.0\% vs 6.5\%).
\begin{table*}[htbp]
    \centering
    \begin{tabular}{r|l|l|l|l}
    \toprule
        Data Split &LF Enc. WM& LF Enc. SM & RN Mod. WM & RN Mod. SM \\
        \hline
        0.01 – 0.99  & 9.0 $\pm$ 0.2  & 4.9 $\pm$ 0.1  & 6.5 $\pm$ 0.1  & \textbf{74.5 $\pm$ 0.1} \\
        0.03 – 0.97  & 23.7 $\pm$ 0.2 & 23.1 $\pm$ 0.7 & 20.9 $\pm$ 0.3 & \textbf{74.3 $\pm$ 0.1} \\
        0.1 – 0.9    & 46.7 $\pm$ 0.2 & 50.0 $\pm$ 0.6 & 46.8 $\pm$ 0.4 & \textbf{73.9 $\pm$ 0.1} \\
        0.33 – 0.67  & 60.7 $\pm$ 0.1 & 65.4 $\pm$ 0.2 & 64.9 $\pm$ 0.2 & \textbf{71.6 $\pm$ 0.1} \\
        0.5 – 0.5    & 62.9 $\pm$ 0.3 & 62.8 $\pm$ 0.2 & 69.1 $\pm$ 0.1 & \textbf{69.2 $\pm$ 0.1} \\
    \bottomrule
    \end{tabular}
    \caption{Performance of unimodal subnetworks within a late-fusion classifier (LF Enc. WM, SM) and independently trained unimodal ResNets (ResNet Mod. WM, SM). All models trained on the synthetic version of the bimodal ImageNet dataset.
    \texttt{WM} and \texttt{SM} denote \texttt{weaker} and \texttt{stronger} modality respectively.
    }
    \label{table:in_2_mods_encoders_vs_resnets}
\end{table*}

\begin{figure}[htbp]
\centering
\begin{subfigure}[t]{0.48\linewidth}
    \centering
    \includegraphics[width=1.\linewidth]{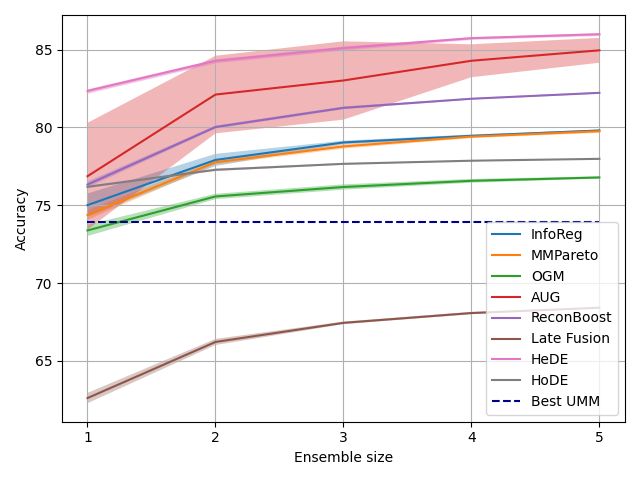}
    \caption{ImageNet-1K (0.1 -- 0.9).}
    \label{fig:acc_vs_ens_size_all_methods_imagenet_0.1_0.9}
\end{subfigure}
\hfill
\begin{subfigure}[t]{0.48\linewidth}
    \centering
    \includegraphics[width=\linewidth]{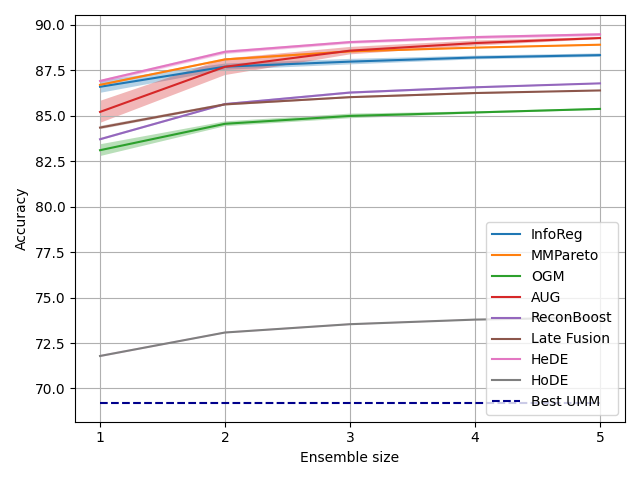}
    \caption{ImageNet-1K (0.5 -- 0.5).}
    \label{fig:acc_vs_ens_size_all_methods_imagenet_0.5_0.5}
\end{subfigure}
\caption{Under milder modality imbalance, HeDEs significantly outperform state-of-the-art late-fusion methods designed to handle modality imbalance.}
\label{fig:acc_vs_ens_size_imagenet_two_example}
\end{figure}

In Figure~\ref{fig:acc_vs_ens_size_imagenet_two_example} we show how accuracy of deep ensembles and imbalance-aware late-fusion methods changes with the ensemble size.
The gaps in performance between HeDEs and the second-best methods are narrower than in the case of the bimodal CIFAR-100 (see Fig.~\ref{fig:acc_vs_ens_size_cifar_two_example}).

\paragraph{Trimodal setup.} We extend our analysis to three modalities on synthetic ImageNet-1K. Table~\ref{tab:trimodal_results} compares accuracies for the naive late-fusion network, homogeneous ensembles of each modality, and heterogeneous ensembles of size 3 across ten modality splits.

As in the bimodal case, HeDEs outperform late-fusion networks and HoDEs when modality imbalance is not extreme. For the most imbalanced splits of the form $(0.01\text{--}r_2\text{--}r_3)$, HoDEs trained on the strongest modality are preferable. Notably, unlike the bimodal case, the late-fusion network's performance does not consistently improve as one of the weaker modalities become stronger. For example, comparing (0.01--0.03--0.96) at 12.4\% with (0.01--0.3--0.69) at 10.7\%, we see that redistributing data from the strongest to the second modality actually hurts the late-fusion network.
\begin{table}[htbp]
    \centering
    \begin{tabular}{l|c|c|c|c|c}
    \toprule
        Data split & Late Fusion & HoDE (Mod. 1) & HoDE (Mod. 2) & HoDE (Mod. 3) & HeDE \\
        \hline
        0.01–0.01–0.99 & 14.3 $\pm$ 0.6 & 7.1 $\pm$ 0.1 & 6.6 $\pm$ 0.1 & \textbf{77.3 $\pm$ 0.1} & 55.9 $\pm$ 0.3 \\
        0.01–0.03–0.96 & 12.4 $\pm$ 0.5 & 7.1 $\pm$ 0.1 & 23.0 $\pm$ 0.1 & \textbf{77.2 $\pm$ 0.1} & 66.3 $\pm$ 0.2 \\
        0.01–0.1–0.89  & 13.5 $\pm$ 0.5 & 7.1 $\pm$ 0.1 & 50.5 $\pm$ 0.1 & \textbf{76.6 $\pm$ 0.1} & 75.9 $\pm$ 0.2 \\
        0.01–0.3–0.69  & 10.7 $\pm$ 0.3 & 7.1 $\pm$ 0.1 & 67.4 $\pm$ 0.1 & 75.1 $\pm$ 0.1 & \textbf{80.9 $\pm$ 0.1} \\
        0.03–0.03–0.94 & 39.6 $\pm$ 0.3 & 23.1 $\pm$ 0.2 & 23.3 $\pm$ 0.2 & \textbf{77.0 $\pm$ 0.1} & 74.5 $\pm$ 0.2 \\
        0.03–0.1–0.84  & 38.9 $\pm$ 0.9 & 23.1 $\pm$ 0.2 & 50.4 $\pm$ 0.1 & 76.6 $\pm$ 0.1 & \textbf{82.0 $\pm$ 0.1} \\
        0.03–0.3–0.67  & 37.7 $\pm$ 0.8 & 23.1 $\pm$ 0.2 & 67.5 $\pm$ 0.1 & 74.9 $\pm$ 0.1 & \textbf{85.8 $\pm$ 0.1} \\
        0.1–0.1–0.8    & 72.3 $\pm$ 0.5 & 50.6 $\pm$ 0.1 & 50.7 $\pm$ 0.1 & 76.1 $\pm$ 0.1 & \textbf{87.3 $\pm$ 0.1} \\
        0.1–0.3–0.6    & 75.8 $\pm$ 0.6 & 50.5 $\pm$ 0.2 & 67.6 $\pm$ 0.1 & 74.1 $\pm$ 0.1 & \textbf{89.7 $\pm$ 0.1} \\
        0.33–0.33–0.34 & 87.1 $\pm$ 0.1 & 68.6 $\pm$ 0.1 & 69.1 $\pm$ 0.1 & 66.8 $\pm$ 0.1 & \textbf{91.3 $\pm$ 0.1} \\
    \bottomrule
    \end{tabular}
    \caption{Performance of the naive late-fusion classifier, homogeneous deep ensembles for each modality, and heterogeneous deep ensembles of size 3. All models trained on the synthetic trimodal ImageNet-1K dataset.}
    \label{tab:trimodal_results}
\end{table}

To understand this behavior, we analyze individual encoders within the late-fusion network (Table~\ref{tab:encoder_results}). For splits of the form $(0.01\text{--}r_2\text{--}r_3)$, all three encoders perform poorly regardless of how much data their modality has access to. 
This suggests that the late-fusion network reaches zero training loss before any encoder, including the one trained on the strongest modality, has extracted discriminative features. 
For less imbalanced splits, encoder performance correlates with modality strength as expected: the encoder with access to more data achieves higher accuracy.
\begin{table}[htbp]
    \centering
    \begin{tabular}{l|c|c|c|c|c|c}
    \toprule
        Data split & LF Enc. 1 & LF Enc. 2 & LF Enc. 3 & RN Mod. 1 & RN Mod. 2 & RN Mod. 3 \\
        \hline
        0.01–0.01–0.99 & 6.7 $\pm$ 0.2 & 6.4 $\pm$ 0.3 & 2.6 $\pm$ 0.3 & 6.5 $\pm$ 0.1 & 6.1 $\pm$ 0.1 & \textbf{74.4 $\pm$ 0.2} \\
        0.01–0.03–0.96 & 6.1 $\pm$ 0.3 & 8.7 $\pm$ 0.2 & 5.2 $\pm$ 0.2 & 6.5 $\pm$ 0.1 & 20.6 $\pm$ 0.2 & \textbf{74.3 $\pm$ 0.2} \\
        0.01–0.1–0.89  & 6.8 $\pm$ 0.3 & 7.8 $\pm$ 0.5 & 7.4 $\pm$ 0.6 & 6.5 $\pm$ 0.1 & 46.8 $\pm$ 0.2 & \textbf{73.8 $\pm$ 0.1} \\
        0.01–0.3–0.69  & 6.7 $\pm$ 0.3 & 6.0 $\pm$ 0.2 & 6.1 $\pm$ 0.2 & 6.5 $\pm$ 0.1 & 63.8 $\pm$ 0.1 & \textbf{71.9 $\pm$ 0.1} \\
        0.03–0.03–0.94 & 19.1 $\pm$ 0.4 & 18.3 $\pm$ 0.4 & 15.1 $\pm$ 0.3 & 20.8 $\pm$ 0.3 & 20.9 $\pm$ 0.3 & \textbf{74.1 $\pm$ 0.1} \\
        0.03–0.1–0.84  & 16.5 $\pm$ 0.3 & 26.5 $\pm$ 0.5 & 25.5 $\pm$ 0.3 & 20.8 $\pm$ 0.3 & 46.7 $\pm$ 0.1 & \textbf{73.6 $\pm$ 0.1} \\
        0.03–0.3–0.67  & 16.2 $\pm$ 0.2 & 26.2 $\pm$ 0.6 & 26.1 $\pm$ 0.3 & 20.8 $\pm$ 0.3 & 63.8 $\pm$ 0.1 & \textbf{71.7 $\pm$ 0.1} \\
        0.1–0.1–0.8    & 41.9 $\pm$ 0.9 & 42.3 $\pm$ 0.3 & 42.4 $\pm$ 0.3 & 46.8 $\pm$ 0.4 & 47.1 $\pm$ 0.3 & \textbf{73.0 $\pm$ 0.2} \\
        0.1–0.3–0.6    & 34.6 $\pm$ 0.9 & 51.8 $\pm$ 0.5 & 51.6 $\pm$ 0.5 & 46.8 $\pm$ 0.4 & 63.9 $\pm$ 0.1 & \textbf{70.8 $\pm$ 0.2} \\
        0.33–0.33–0.34 & 52.9 $\pm$ 0.2 & 52.9 $\pm$ 0.3 & 52.5 $\pm$ 0.3 & 64.9 $\pm$ 0.2 & \textbf{65.4 $\pm$ 0.1} & 65.1 $\pm$ 0.1 \\
    \bottomrule
    \end{tabular}
    \caption{Performance of unimodal subnetworks within the trimodal late-fusion classifier (LF Enc. 1, 2, 3) and independently trained unimodal ResNets (RN Mod. 1, 2, 3). All models trained on the synthetic trimodal ImageNet-1K dataset.}
    \label{tab:encoder_results}
\end{table}

\subsection{CREMA-D}
\label{appendix_add_results_crema}
Many existing approaches train and evaluate their late-fusion methods on CREMA-D using data splits that suffer from data leakage.
CREMA-D comprises video recordings on which various actors utter phrases with certain emotions.
The existing works do not ensure that different actors from the original CREMA-D dataset appear exclusively in either the training or test set. 
As a result, videos featuring the same actor can appear in both partitions, allowing models to exploit actor-specific features (such as voice characteristics or facial appearance) rather than learning generalizable multimodal representations. 
This leads to artificially inflated performance metrics that do not reflect true generalization ability.

To illustrate the impact of data leakage, we compare the performance of existing approaches on both the leaked split (as used in prior work) and our clean split in Table~\ref{tab:leakage_comparison}. 
As the results show, all methods exhibit a significant drop in accuracy when evaluated on the clean split, highlighting the extent to which previous results were inflated by data leakage.
\begin{table}[htbp]
    \centering
    \begin{tabular}{l|c|c|c}
    \toprule
        Method & Data Leak & No Data Leak & $\Delta$ \\
        \hline
        OGM & 63.4 & 56.7 & 6.7 \\
        MMPareto & 69.1 & 61.2 & 7.9\\
        InfoReg & 69.0 & 59.7 & 9.3 \\
        ReconBoost & 75.4 & 61.7 & 13.7\\
        AUG & 80.9 & 65.0 & 15.9 \\
        LFM & 80.1 & 63.2 & 16.9\\
    \bottomrule
    \end{tabular}
    \caption{Performance comparison of existing late-fusion methods on CREMA-D with and without data leakage. All methods show a significant drop in accuracy when evaluated on the clean split without data leakage.}
    \label{tab:leakage_comparison}
\end{table}

In our experiments, we address this issue by constructing a clean data split in which actors are strictly separated between the training and test sets. 
Each actor appears in only one partition, ensuring that the model cannot rely on actor identity to improve predictions. 
This setup provides a more realistic and rigorous evaluation of multimodal classification methods.

\begin{figure}[htbp]
    \centering
    \includegraphics[width=0.6\linewidth]{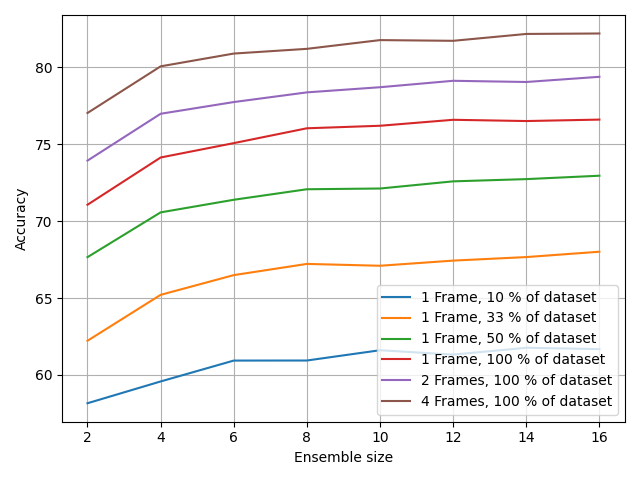}
    \caption{Performance of heterogeneous deep ensembles as a function of ensemble size. We construct different ensembles by keeping the audio modality fixed while varying the data available to the video modality and the number of input frames, as described in the main text.}
    \label{fig:crema_hede_scale_up}
\end{figure}

As discussed in Section~\ref{sec:scaling_laws}, we can make both audio and video modalities weaker by reducing the amount of training data available to the models or make the video modality stronger by using more frames to classify samples.
Figure~\ref{fig:crema_hede_scale_up} shows how the performance of heterogeneous deep ensembles changes as we vary the strength of the video modality while keeping the strength of the audio modality fixed.


\newpage
\subsection{Sarcasm}
\label{appendix_add_results_sarcasm}
As mentioned in Section~\ref{sec:results}, we use a pre-trained BERT-Base model for the textual modality and ResNet-50 pre-trained on the ImageNet-1K dataset for the visual modality.
Next, we fine-tune the textual and visual models independently on the corresponding modalities to form a heterogeneous deep ensemble.
To assess the quality of late-fusion methods, we concatenate the modality-specific features and apply imbalance-aware strategies while fine-tuning.
Both encoders and the linear classification layer are fine-tuned jointly.

\begin{figure}[htbp]
    \centering
    \includegraphics[width=0.6\linewidth]{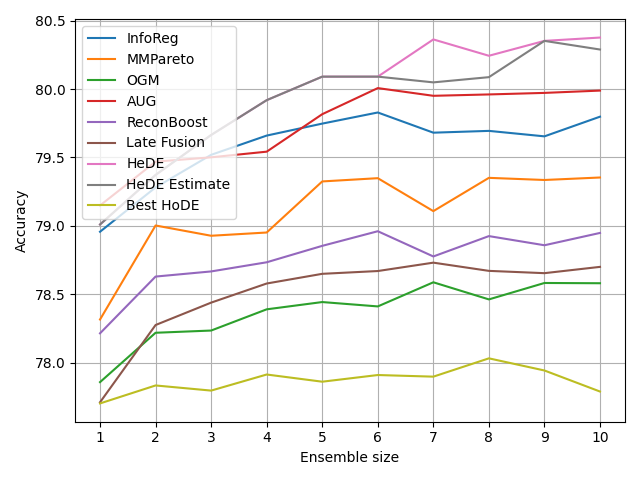}
    \caption{Performance of deep ensembles and late-fusion approaches on the Sarcasm dataset. }
    \label{fig:sarcasm_fine_tune_pre_trained}
\end{figure}

Figure~\ref{fig:sarcasm_fine_tune_pre_trained} shows how accuracy of different methods change with the ensemble size.
For fair parameter count, deep ensembles use twice as many models as late-fusion baselines.
For the Sarcasm dataset, ensembling the fine-tuned textual encoders yield only marginal improvement.
However, multimodal approaches show more robust improvement when scaled up, and heterogeneous deep ensembles being the best, highlighting that ensembles can use information from both modalities more efficiently.

\subsection{Heuristics}
\label{subsec:appendxi_heuristic}

In this subsection, we validate that our heuristic produces reliable
allocation estimates across synthetic and real-world datasets for
modalities of varying strength.
\begin{figure}[htbp]
\centering
\begin{subfigure}[t]{0.48\linewidth}
    \centering
    \includegraphics[width=\linewidth]{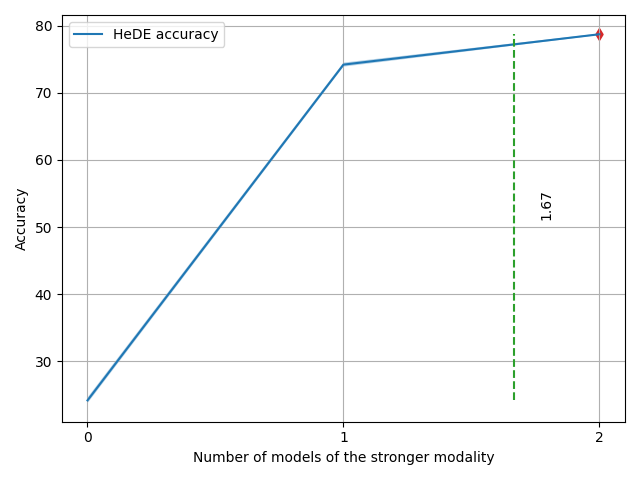}
    \caption{CIFAR-100 (0.05 -- 0.95). Ensemble size 2.}
    \label{c_100_0.05_0.95_size_2}
\end{subfigure}
\hfill
\begin{subfigure}[t]{0.48\linewidth}
    \centering
    \includegraphics[width=\linewidth]{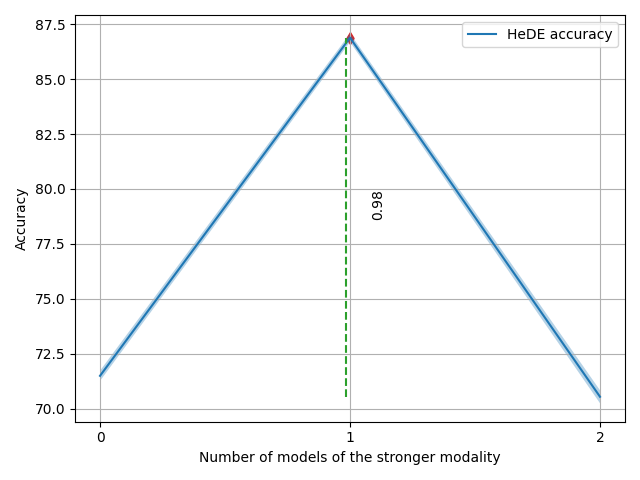}
    \caption{CIFAR-100 (0.5 -- 0.5). Ensemble size 2.}
    \label{c_100_0.5_0.5_size_2}
\end{subfigure}

\vspace{0.5em}

\begin{subfigure}[t]{0.48\linewidth}
    \centering
    \includegraphics[width=\linewidth]{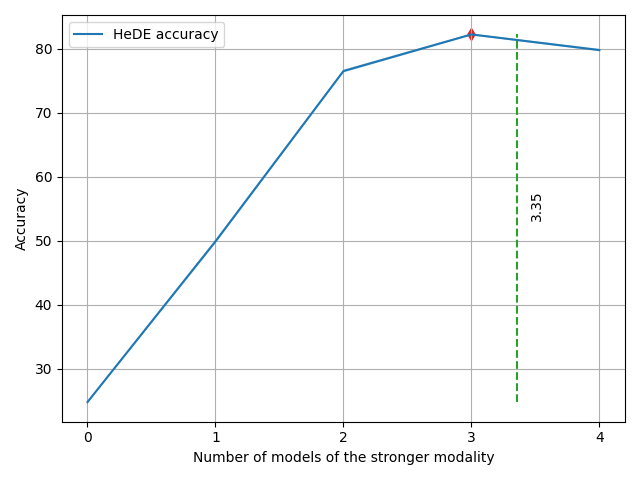}
    \caption{CIFAR-100 (0.05 -- 0.95). Ensemble size 4.}
    \label{c_100_0.05_0.95_size_4}
\end{subfigure}
\hfill
\begin{subfigure}[t]{0.48\linewidth}
    \centering
    \includegraphics[width=\linewidth]{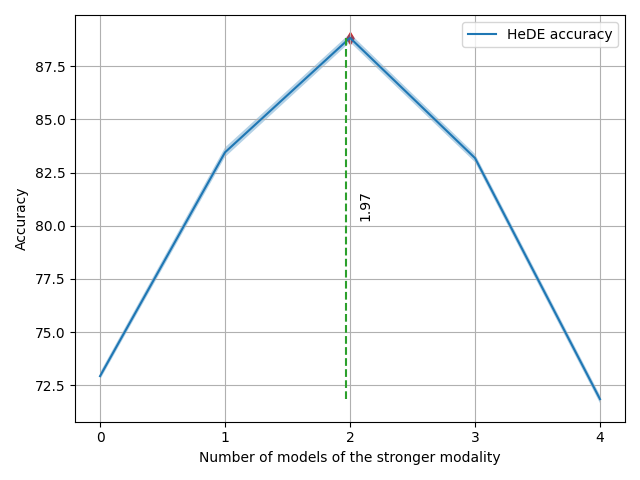}
    \caption{CIFAR-100 (0.5 -- 0.5). Ensemble size 4.}
    \label{c_100_0.5_0.5_size_4}
\end{subfigure}
\caption{Bimodal CIFAR-100 dataset. Heuristics correctly predicts switching to heterogeneous deep ensembles for extremely imbalanced modalities when scaling up the ensemble size. In all subfigures, the green dashed line is the number of models trained on a stronger modality predicted by our heuristic.}
\label{fig:cifar_scale_up_heuristic_works}
\end{figure}

\begin{figure}[htbp]
\centering
\begin{subfigure}[t]{0.48\linewidth}
    \centering
    \includegraphics[width=\linewidth]{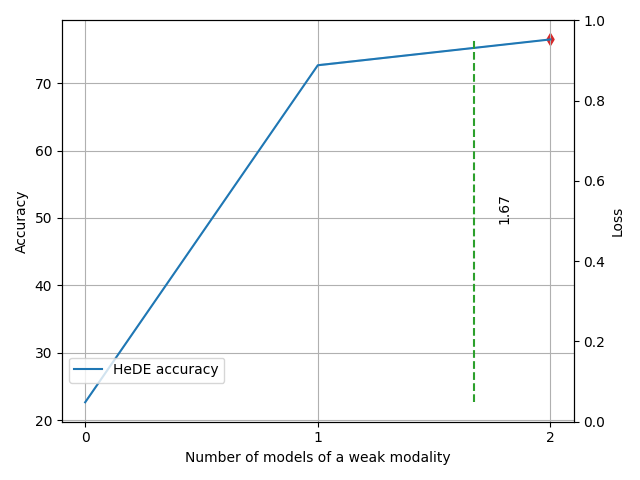}
    \caption{ImageNet-1K (0.03 -- 0.97). Ensemble size 2.}
    \label{in_1k_0.03_0.97_size_2}
\end{subfigure}
\hfill
\begin{subfigure}[t]{0.48\linewidth}
    \centering
    \includegraphics[width=\linewidth]{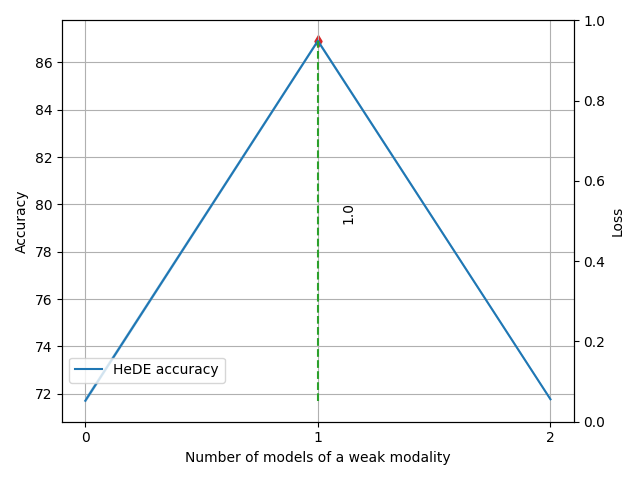}
    \caption{ImageNet-1K (0.5 -- 0.5). Ensemble size 2.}
    \label{in_1k_0.5_0.5_size_2}
\end{subfigure}

\vspace{0.5em}

\begin{subfigure}[t]{0.48\linewidth}
    \centering
    \includegraphics[width=\linewidth]{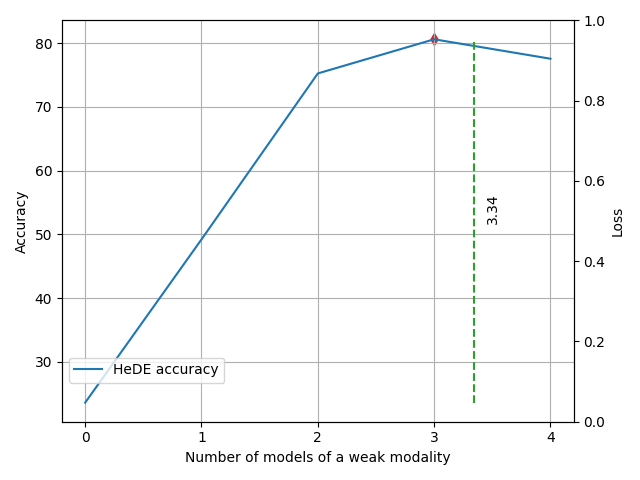}
    \caption{ImageNet-1K (0.03 -- 0.97). Ensemble size 4.}
    \label{in_1k_0.03_0.97_size_4}
\end{subfigure}
\hfill
\begin{subfigure}[t]{0.48\linewidth}
    \centering
    \includegraphics[width=\linewidth]{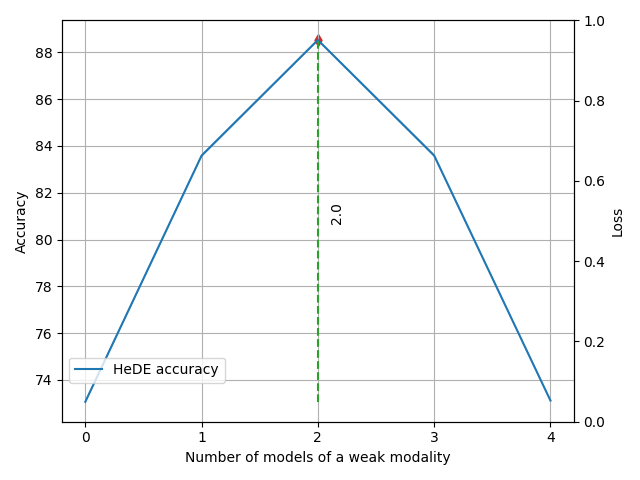}
    \caption{ImageNet-1K (0.5 -- 0.5). Ensemble size 4.}
    \label{in_1k_0.5_0.5_size_4}
\end{subfigure}
\caption{Bimodal ImageNet-1K dataset. Heuristics correctly predicts switching to heterogeneous deep ensembles for extremely imbalanced modalities when scaling up the ensemble size. In all subfigures, the green dashed line is the number of models trained on a stronger modality predicted by our heuristic.}
\label{fig:in_scale_up_heuristic_works}
\end{figure}

\begin{figure}[htbp]
\centering
\begin{subfigure}[t]{0.48\linewidth}
    \centering
    \includegraphics[width=\linewidth]{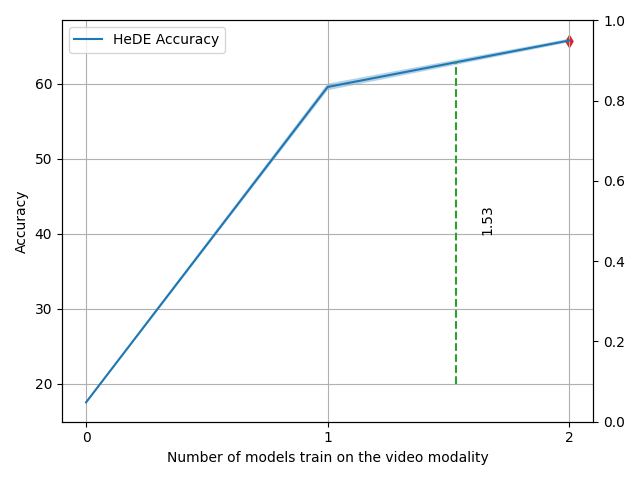}
    \caption{Kinetics-400. Ensemble size 2.}
    \label{k400_size_2}
\end{subfigure}
\hfill
\begin{subfigure}[t]{0.48\linewidth}
    \centering
    \includegraphics[width=\linewidth]{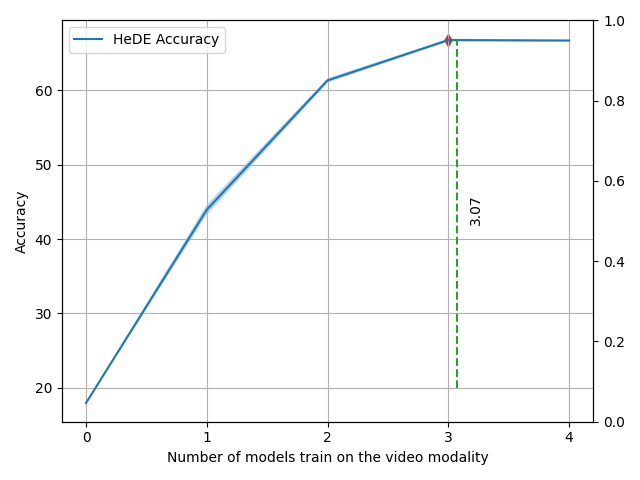}
    \caption{Kinetics-400. Ensemble size 4.}
    \label{k400_size_4}
\end{subfigure}
\caption{Kinetics-400. Heuristics correctly predicts switching to heterogeneous deep ensembles for extremely imbalanced modalities when scaling up the ensemble size. In all subfigures, the green dashed line is the number of models trained on a stronger modality predicted by our heuristic.}
\label{fig:k400_scale_up_heuristic_works}
\end{figure}

We begin with the bimodal case. Figs.~\ref{fig:cifar_scale_up_heuristic_works}
and~\ref{fig:in_scale_up_heuristic_works} compare heterogeneous and
homogeneous deep ensembles at sizes 2 and 4 on synthetic CIFAR-100
and ImageNet-1K, respectively. In each subfigure, the leftmost and
rightmost points correspond to homogeneous ensembles of the weaker
and stronger modality; the dashed green line indicates the number of
stronger-modality models predicted by our heuristic. For ensemble
size 2 under extreme imbalance, the heuristic correctly predicts
that a homogeneous ensemble is preferable; at size 4, it predicts
the transition to a heterogeneous ensemble, consistent with the
brute-force optimum. Fig.~\ref{fig:k400_scale_up_heuristic_works} shows
analogous results on Kinetics-400, confirming the same pattern on a
real-world dataset.

\begin{figure}[htbp]
\centering
\begin{subfigure}[t]{0.48\linewidth}
    \centering
    \includegraphics[width=\linewidth]{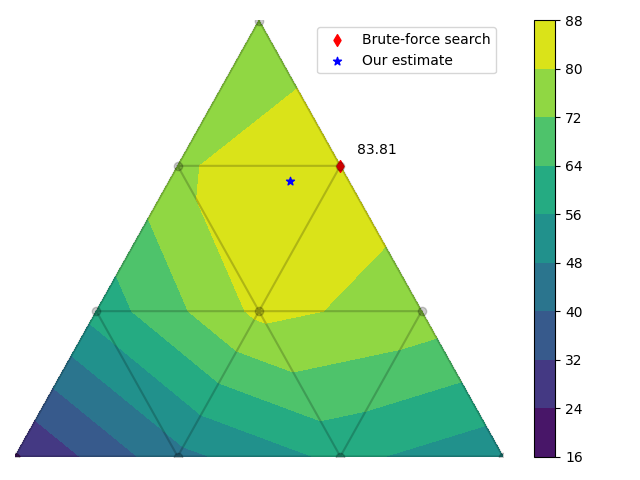}
    \caption{Trimodal ImageNet-1K dataset (0.03 -- 0.1 -- 0.87).}
    \label{in_trimodal_3_10_87_size_3}
\end{subfigure}
\hfill
\begin{subfigure}[t]{0.48\linewidth}
    \centering
    \includegraphics[width=\linewidth]{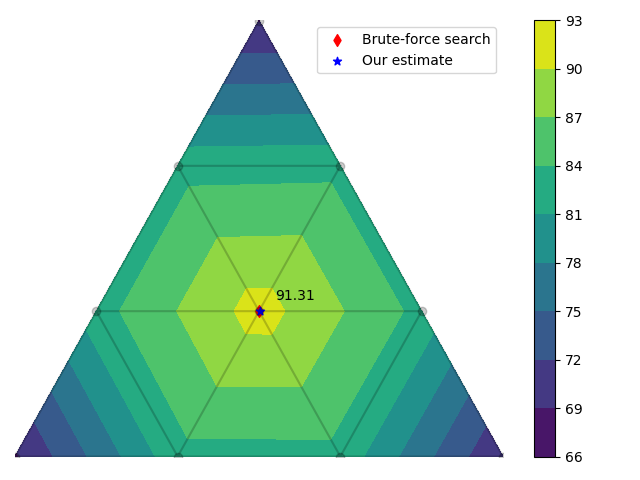}
    \caption{Trimodal ImageNet-1K dataset (0.33 -- 0.33 -- 0.34).}
    \label{in_trimodal_33_33_33_size_3}
\end{subfigure}

\vspace{0.5em}

\begin{subfigure}[t]{0.48\linewidth}
    \centering
    \includegraphics[width=\linewidth]{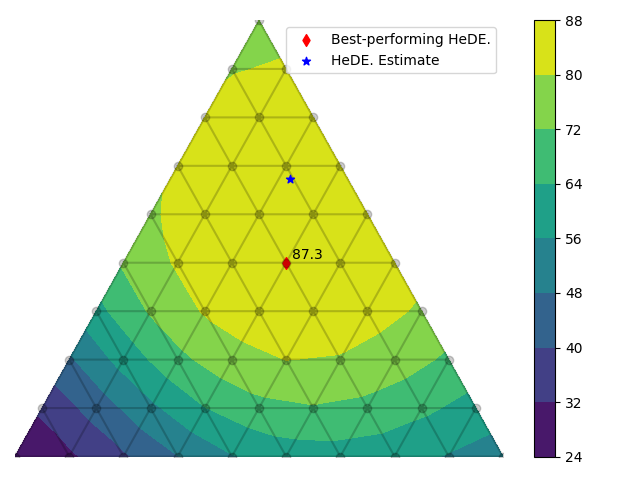}
    \caption{Trimodal ImageNet-1K dataset (0.03 -- 0.1 -- 0.87).}
    \label{in_trimodal_3_10_87_size_9}
\end{subfigure}
\hfill
\begin{subfigure}[t]{0.48\linewidth}
    \centering
    \includegraphics[width=\linewidth]{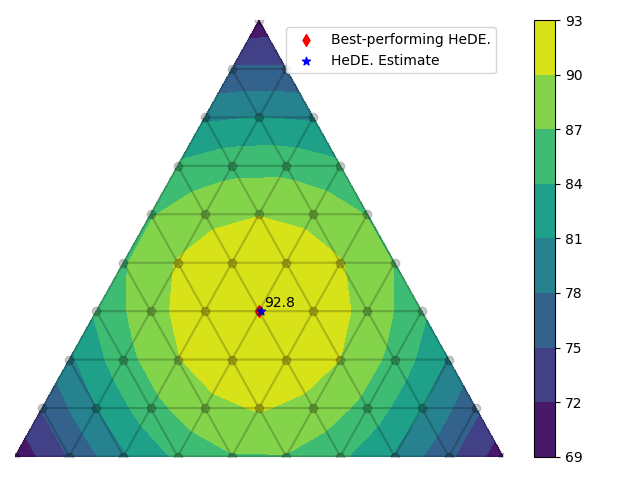}
    \caption{Trimodal ImageNet-1K dataset (0.33 -- 0.33 -- 0.34).}
    \label{in_trimodal_33_33_33_size_9}
\end{subfigure}
\caption{Trimodal ImageNet-1K datasets. Brute-force solution is a red diamond. Our heuristic is a blue star.}
\label{fig:in_trimodal}
\end{figure}

We further validate the heuristic on trimodal synthetic ImageNet-1K
(Fig.~\ref{fig:in_trimodal}). Even under extreme modality
imbalance, the heuristic predicts near-optimal allocations across
all configurations tested.

Finally, we evaluate the heuristic on CREMA-D with modalities of
varying strength. As discussed in Section~\ref{sec:scaling_laws},
we can weaken a modality by reducing the amount of training data or
strengthen it by using a more powerful encoder.
Table~\ref{tab:crema_opt_vs_heuristic_modalities_of_different_strength}
shows the heuristic's predictions across these configurations: for
the audio modality, we use either 33\% or the full dataset; for the
video modality, we use the same data fractions and additionally vary
the encoder by increasing the number of input frames. In most
configurations, rounding the heuristic's prediction yields the
brute-force-optimal allocation; in the remaining cases, the error
does not exceed one model. This result also validates the heuristic
in a semi-synthetic setting: starting from the instance-aligned
CREMA-D dataset and producing modality-imbalanced variants via our
construction from Section~\ref{sec:motivation}.


\begin{table}[]
    \centering
    \begin{tabular}{c|c|c|c|c|c|c|c}
    \toprule
    Ensemble Size & Audio DF & Video DF & Video NoF & Optimum & Heuristic & $|\Delta|$ & Correct \\
    \hline
    \multirow{6}{*}{2} & 0.33 & 0.33 & 1 & 1 & 0.91 & 0.09 & \cmark \\
    & 0.33 & 1.0 & 1 & 1 & 1.26 & 0.26 & \cmark \\
    & 0.33 & 1.0 & 4 & 1 & 1.31 & 0.31 & \cmark \\
    & 1.0 & 0.33 & 1 & 1 & 0.74 & 0.26 & \cmark \\
    & 1.0 & 1.0 & 1 & 1 & 1.09 & 0.09 & \cmark \\
    & 1.0 & 1.0 & 4 & 1 & 1.15 & 0.15 & \cmark \\
    \hline
    \multirow{6}{*}{5} & 0.33 & 0.33 & 1 & 3 & 2.27 & 0.73 & \xmark \\
    & 0.33 & 1.0 & 1 & 4 & 3.14 & 0.86 & \xmark \\
    & 0.33 & 1.0 & 4 & 3 & 3.29 & 0.29 & \cmark \\
    & 1.0 & 0.33 & 1 & 2 & 1.85 & 0.15 & \cmark \\
    & 1.0 & 1.0 & 1 & 3 & 2.73 & 0.27 & \cmark \\
    & 1.0 & 1.0 & 4 & 3 & 2.88 & 0.12 & \cmark \\
    \hline
    \multirow{6}{*}{10}  & 0.33 & 0.33 & 1 & 6 & 4.54 & 1.46 & \xmark \\
    & 0.33 & 1.0 & 1 & 7 & 6.28 & 0.72 & \xmark \\
    & 0.33 & 1.0 & 4 & 6 & 6.57 & 0.57 & \xmark \\
    & 1.0 & 0.33 & 1 & 4 & 3.71 & 0.29 & \cmark \\
    & 1.0 & 1.0 & 1 & 6 & 5.45 & 0.55 & \xmark \\
    & 1.0 & 1.0 & 4 & 5 & 5.76 & 0.76 & \xmark \\
    \bottomrule
    \end{tabular}
    \caption{Heuristic vs.\ brute-force-optimal allocation of video
    models in heterogeneous deep ensembles on CREMA-D.
    \texttt{DF}: fraction of the training data used to train each
    unimodal model. \texttt{NoF}: number of frames used by the
    unimodal video model. \texttt{Optimum}: brute-force-optimal
    number of video models. \texttt{Heuristic}: number predicted
    by Eq.~\eqref{eq:heuristics}. $|\Delta|$: absolute difference
    between the two. \texttt{Correct}: whether rounding the
    heuristic yields the optimum (\cmark) or not (\xmark).}
    \label{tab:crema_opt_vs_heuristic_modalities_of_different_strength}
\end{table}

\subsection{Noise corruption}
\label{subsec:appendex_noise_corruption}
The setup suggested in Section~\ref{sec:synthetic_data} allows us to vary the amount of data available to each of the modalities.
Reducing the number of samples available to a modality results in lower predictive strength and faster convergence; conversely, increasing the data available improves generalization but slows convergence.
On the other hand, we can force a modality to converge longer and simultaneously reducing its predictive strength by adding noise to input samples.

We conducted additional experiments on the synthetic CIFAR-100 (0.5 -- 0.5) dataset to address this regime. 
Starting from two modalities of equal strength, we inject fixed per-image Gaussian noise with a given standard deviation into one of them. 
This makes the corrupted modality weaker and causes it to converge more slowly. 
Table~\ref{tab:syntheric_noisy_dataset_results} shows results with the clean setting in column "Noise 0.0". 
"Vanilla LF" denotes the vanilla late-fusion baseline. 
The number of stronger-modality models is determined by Formula \ref{eq:heuristics} in Section~\ref{subsec:number_of_networks}, which yields a real-valued estimate that we round to the nearest integer for the actual ensemble configuration. 
Under mild noise, HeDE performance degrades gracefully but still outperforms both LF and HoDE. When noise becomes substantial, our heuristic correctly indicates that HoDE is preferable — consistent with our findings on highly imbalanced modalities in Sections~\ref{subsec:results_synthetic} and \ref{subsec:results_on_real_data}.
\begin{table}[htbp]
    \centering
    \begin{tabular}{c|c|c|c}
    \toprule
        Model & Noise 0.0 (baseline) & Noise 0.1 & Noise 2.0 \\
        \hline
        Vanilla LF & 82.5 & 79.6 & 70.2 \\
        HeDE(2) & 87.3 & 83.2 & 57.3 \\
        HoDE(2) & 71.7 & 71.7 & 71.7 \\
        Weaker UM & 68.7 & 57.8 & 7.0 \\
        Heuristic  & 0.99 & 1.17 & 1.69 \\
    \bottomrule
    \end{tabular}
    \caption{
    Synthetic bimodal CIFAR-100 (0.5 -- 0.5) dataset. 
    Injecting Gaussian noise to input images makes the noisy modality weaker and forces to converge longer which results in lower late-fusion accuracy.
    In this setup, our heuristics correctly predicts when homogeneous ensembles are preferable to heterogeneous counterparts.
    }
    \label{tab:syntheric_noisy_dataset_results}
\end{table}

%% file: content/section_appendix_b_training_details.tex
\paragraph{ResNets on CIFAR-100 and ImageNet}
On CIFAR-100, ResNet-18 models are trained with SGD (momentum 0.9, weight decay $5\times10^{-4}$),
an initial learning rate of 0.1 decayed linearly to $10^{-3}$ over 100 epochs, and a batch
size of 256. Training images are augmented with random cropping (32x32 with padding 4,
reflect mode) and random horizontal flipping, then normalised with the CIFAR-100 channel-wise mean
[0.505, 0.485, 0.439] and standard deviation [0.266, 0.255, 0.275].

On ImageNet, ResNet-50 models are trained with SGD (momentum 0.9, weight decay $10^{-4}$),
an initial learning rate of 0.1 decayed with cosine annealing over 100 epochs, and a
batch size of 256. Training images use RandomResizedCrop(224) and random horizontal
flipping, normalised with ImageNet statistics (mean [0.485, 0.456, 0.406], std [0.229,
0.224, 0.225]).

In both cases, multimodal networks (those with two encoder bodies) use the same
hyperparameters as their unimodal counterparts.

\paragraph{ViTs on ImageNet}
ViT-B/16 models are trained with AdamW (beta1=0.9, beta2=0.999, weight decay 0.1), an
initial learning rate of $10^{-3}$ warmed up linearly over 5 epochs and then decayed with
cosine annealing over 300 epochs total. Batch size is 128. Training uses
RandomResizedCrop(224) with bicubic interpolation, random horizontal flipping,
ColorJitter (brightness, contrast, saturation each 0.4), RandAugment (2 operations,
magnitude 9), and RandomErasing (p=0.25). Mixup (alpha 0.8) and CutMix (alpha 1.0) are
applied, along with label smoothing (epsilon=0.1), dropout (0.1), and stochastic depth
(drop path rate 0.1). An exponential moving average of the model weights is maintained
with decay 0.9999, and gradients are clipped to a maximum norm of 1.0.

Multimodal ViT configurations use the same training procedure, extended with cross-
attention blocks between encoder bodies; no changes to the optimiser or augmentation
pipeline are made for the multimodal case.

\paragraph{ResNets on CREMA-D}
CREMA-D contains video and audio (mel spectrogram) modalities. ResNet-18 models are
used as encoders for both. For all baselines, audio is represented as log-magnitude spectrograms computed via short-time Fourier transform (nfft=512, hoplength=353) on waveforms resampled to 22,050 Hz, without further normalization; MMPareto instead uses 128-bin Kaldi mel-filterbank features normalized by global mean and standard deviation. 
Video clips are sampled as single frames, randomly resized and cropped to 224x224 during
training and centre-cropped at test time, with random horizontal flipping during
training. Video clips are normalised with mean [0.485, 0.456, 0.406] and standard
deviation [0.229, 0.224, 0.225]. All encoders are trained from random initialisation.

Audio unimodal models are trained with SGD (momentum 0.9, weight decay $10^{-4}$), batch size 32, and cosine annealing LR schedule starting at 0.1 for 64 epochs. 
Video unimodal models use a higher weight decay ($10^{-3}$) and are trained for 256 epochs with batch size 64. Multimodal models follow the video hyperparameters but is trained on for 64 epochs (weight decay $10^{-3}$, batch size 64). 
The difference in training duration and regularisation
between audio and video unimodal models reflects the different complexities of the two modalities on this dataset.

\paragraph{Networks on Kinetics-400}
Kinetics-400 contains video and audio modalities. Video is encoded with an
R(2+1)D-18 network; audio is encoded with a ResNet-50 operating on log-mel
spectrograms (16 kHz, nfft=512, hoplength=512, 40 mel bands, Slaney norm, HTK scale,
40x100 output, tiled across 3 channels as input to ResNet-50). All
models are trained with SGD (momentum 0.9, weight decay $10^{-4}$), an initial learning
rate of 0.1 decayed with cosine annealing over 50 epochs.

For video, 8 frames are sampled per clip with a temporal dilation of 2. Training
augmentation includes a random resize to 256-320 pixels, random crop to 224x224, and
random horizontal flip, with normalisation using mean/std of [0.45, 0.45, 0.45] and
[0.225, 0.225, 0.225] respectively. Validation uses a single centre crop from 256
pixels, averaged over 10 uniformly sampled clips.

Unimodal video models use batch size 128. Unimodal audio models use a substantially
larger batch size of 1024, reflecting the relatively lower per-sample cost of
spectrogram inputs. Multimodal models fuse video (512-dim) and audio (2048-dim) feature
vectors by concatenation followed by a linear classification head. Multimodal training
uses batch size 128 when pairing R(2+1)D-18 with ResNet-50, and 32 for heavier model
combinations.

\paragraph{Modifications to Baseline Implementations}
All baselines were evaluated on CREMA-D using their official codebases with the
following modifications, each of which resulted in improved performance over the
original configuration. OGM-GE and LFM were used without modification.
LFM results are reported only for CREMA-D. On CIFAR-100 and ImageNet, we were unable to obtain stable training with the authors' original implementation: the training loss diverged across multiple hyperparameter configurations, including those recommended in the original paper. We therefore omit LFM from these datasets.

\begin{itemize}
    \item \textbf{MMPareto:}The number of training epochs was increased from 50 to 100. 
    The step learning rate scheduler (decay by 0.1 at epoch 30) was replaced with a linear schedule decaying from the initial rate to 1\% over the full training duration.
    \item \textbf{InfoReg:} The number of training epochs was increased from 50 to 80. The step learning rate scheduler (decay by 0.1 at epoch 30) was replaced with a linear schedule decaying from the initial rate to 1\% over the full training duration.
    \item\textbf{AUG:} The number of training epochs was reduced from 200 to 100. The step learning rate scheduler (decay by 0.1 at epochs 160 and 180) was replaced with cosine
    annealing over the full training duration.
    \item\textbf{ReconBoost:} The number of training epochs per boosting stage was increased from 4 to 9, and the number of ensemble correction epochs per stage was reduced from 4 to 1. Weight decay was reduced from $5\times10^{-4}$ to $10^{-4}$.
\end{itemize}

%% file: content/section_appendix_e_intermediate_fusion_and_i2m2.tex
\subsection{Implementation details}
\paragraph{Intermediate fusion for CNNs}For CNN encoders (ResNet family) we augment the late-fusion baseline with intermediate-fusion blocks using the Multimodal Transfer Module (MMTM)\cite{Joze_2020_MMTM}. MMTM blocks are inserted between the residual stage layers of the two ResNet bodies (stages 2, 3 and 4) and implement a squeeze-and-excitation style cross-modal attention mechanism.

\paragraph{Intermediate fusion for ViTs}
For ViT encoders we replace MMTM with a bidirectional cross-attention layer (ViTCrossAttention) inserted at configurable transformer block boundaries.
ViT blocks are grouped into four macro-layers (three blocks each for a 12-block ViT), and cross-attention is optionally applied after macro-layers 2, 3 and 4.

Let $x_1$ in $\mathbb{R}^{B \times N_1 \times D_1}$ and $x_2$ in $\mathbb{R}^{B \times N_2 \times D_2}$ be the token sequences of the two ViT streams, where $N_i$ is the number of tokens (patches + class token) and $D_i$ is the embedding dimension. 
Because the two encoders may differ in embedding dimension (heterogeneous ViTs), both
sequences are first projected to a common dimension $D_{proj}$ = $\max(D_1, D_2)$ via learned linear projections.

Inputs are layer-normalised before the cross-attention computation (Pre-LN), which is critical for stable training. 
Using post-layer normalisation in this setting consistently causes gradient blow-up during
training, regardless of the initial learning rate, warmup schedule, or degree of modality imbalance between the two encoders.

Two independent MultiheadAttention (MHA) modules are used, one per direction:
\begin{itemize}
    \item \textbf{Modality 1 attends to modality 2:}
    \begin{gather*}
        x_{1,\text{cross}} = MHA_1(Q=proj_1(LN_1(x_1)),
                          K=proj_2(LN_2(x_2)),
                          V=proj_2(LN_2(x_2))),
    \end{gather*}
    \item \textbf{Modality 2 attends to modality 1:}
    \begin{gather*}
        x_{2,\text{cross}} = MHA_2(Q=proj_2(LN_2(x_2)),
                          K=proj_1(LN_1(x_1)),
                          V=proj_1(LN_1(x_1))).
    \end{gather*}
\end{itemize}
Each MHA uses h=8 heads by default. The attention output is projected back to the original embedding dimension via $out_{proj}$ (identity if dimensions already match).

The cross-modal signal is added to the original token stream through a gated residual connection with learnable scalar gates
\begin{align*}
    x_1' &= x_1 + \operatorname{sigmoid}(g_1) \cdot \operatorname{out\_proj}_1(x_{1,\text{cross}}) \\
    x_2' &= x_2 + \operatorname{sigmoid}(g_2) \cdot \operatorname{out\_proj}_2(x_{2,\text{cross}})
\end{align*}

Here $g_1$ and $g_2$ are scalar parameters initialised to a small negative value (default -5), so that $\operatorname{sigmoid}(g_i) \simeq 0.007$ at the start of training. 
This near-zero initialisation ensures that the cross-attention path contributes negligibly at the beginning, allowing each modality's self-supervised representation to stabilise before cross-modal signals are gradually introduced.


\subsection{Intermediate fusion}
Table~\ref{tab:mmtm_full} extends the comparison from the main text
(Table~\ref{table:mmtm}) to additional datasets and imbalance
levels. Adding MMTM blocks yields marginal or no improvement over
the vanilla late-fusion baseline across all configurations; deep
ensembles of the same parameter count consistently outperform both.

\begin{table}[htbp]
\centering
\begin{tabular}{l|c|c|c|c|c}
\toprule
Dataset & LF & MMTM-1 & MMTM-2 & HoDE(2) & HeDE(2) \\
\midrule
CIFAR-100 (0.05 -- 0.95) & 33.6 & 34.9 & 34.3 & \textbf{78.5} & 74.1 \\
CIFAR-100 (0.1 -- 0.9)   & 55.5 & 52.8 & 52.1 & 77.9 & \textbf{78.4} \\
CIFAR-100 (0.2 -- 0.8)   & 70.0 & 67.9 & 68.6 & 76.8 & \textbf{82.8} \\
CIFAR-100 (0.3 -- 0.7)   & 78.6 & 77.3 & 78.9 & 75.2 & \textbf{85.6} \\
CIFAR-100 (0.5 -- 0.5)   & 82.6 & 83.4 & 83.0 & 71.6 & \textbf{86.9} \\
ImageNet (0.33 -- 0.67)   & 84.3 & 84.0 & 84.1 & 74.1 & \textbf{86.5} \\
ImageNet (0.5 -- 0.5)     & 86.5 & 86.5 & 86.7 & 71.7 & \textbf{86.9} \\
CREMA-D (A-V-1-frame)     & 56.1 & 60.3 & 60.8 & 62.8 & \textbf{71.2} \\
\bottomrule
\end{tabular}
\caption{Top-1 accuracy (\%) of intermediate-fusion classifiers
with MMTM blocks on synthetic and real-world datasets.
\texttt{MMTM-X}: late-fusion network with $X$ MMTM blocks inserted
at progressively earlier layers.
\texttt{HoDE(2)}: homogeneous ensemble of size two trained on the
stronger modality.
\texttt{HeDE(2)}: heterogeneous ensemble of size two.}
\label{tab:mmtm_full}
\end{table}

Table~\ref{tab:cross_attn} reports an analogous comparison using ViT encoders with cross-attention fusion blocks on ImageNet. Although cross-attention blocks can improve over the late-fusion baseline under strong imbalance (e.g., 49.2\% vs.\ 37.3\% at the 0.03–0.97 split), deep ensembles outperform all cross-attention configurations across every imbalance level by a wide margin. Under extreme imbalance (0.01–0.99 and 0.03–0.97), homogeneous ensembles of the stronger modality achieve the highest accuracy; as the imbalance becomes less pronounced (0.1–0.9 and beyond), heterogeneous ensembles take the lead.
\begin{table}[htbp]
\centering
\begin{tabular}{l|c|c|c|c|c|c}
\toprule
Dataset & LF & CA-1 & CA-2 & CA-3 & HoDE(2) & HeDE(2) \\
\midrule
ImageNet (0.01 -- 0.99) & 19.3 & 19.4 & 19.8  & 20.1 & \textbf{81.1}   & 80.0 \\
ImageNet (0.03 -- 0.97) & 37.3 & 49.2 & 43.8  & 43.9 & \textbf{80.7}   & 79.3 \\
ImageNet (0.1 -- 0.9)   & 53.5 & 61.1 & 58.6  & 58.4 & 80.0   & \textbf{83.8} \\
ImageNet (0.33 -- 0.67) & 79.2 & 76.7 & 77.9  & 78.6 & 77.8   & \textbf{87.4} \\
ImageNet (0.5 -- 0.5)   & 84.2 & 82.8 & 84.1  & 84.3 & 76.0   & \textbf{87.5} \\
\bottomrule
\end{tabular}
\caption{Top-1 accuracy (\%) of intermediate-fusion classifiers
with cross-attention blocks on synthetic ImageNet.
\texttt{CA-X}: ViT-based late-fusion network with $X$
cross-attention blocks.}
\label{tab:cross_attn}
\end{table}

\subsection{Hybrid approaches}
\label{subsec:appendxi_hybrid_approahces}
Table~\ref{table:full_de_vs_i2m2} shows the full version of the Table~\ref{table:de_vs_i2m2} from the main text. Here we compare the performance of the best-performing heterogeneous deep ensemble of size four with that of the hybrid I2M2 approach, that averages predictions of both unimodal and the late-fusion multimodal networks.
On CREMA-D, we trained a 2D CNN using a single frame and a 3D CNN that receives four frames as inputs. 
We report accuracies for both configurations.
\begin{table}[htbp]
\centering
\begin{tabular}{l|l|l}
\toprule
    Dataset & HeDE(4) & I2M2 \\
    \hline
    CIFAR-100 (0.05 -- 0.95)    & \textbf{82.0 $\pm$ 0.2} & 74.9 $\pm$ 0.5 \\
    CIFAR-100 (0.1 -- 0.9)      & \textbf{84.0 $\pm$ 0.2} & 80.3 $\pm$ 0.3 \\
    CIFAR-100 (0.2 -- 0.8)      & \textbf{85.7 $\pm$ 0.2} & 85.0 $\pm$ 0.3 \\
    CIFAR-100 (0.3 -- 0.7)      & \textbf{87.7 $\pm$ 0.2} & 87.2 $\pm$ 0.2 \\
    CIFAR-100 (0.5 -- 0.5)      & \textbf{88.9 $\pm$ 0.2} & 87.9 $\pm$ 0.2 \\
    ImageNet-1K (0.01 -- 0.99)   & \textbf{77.7 $\pm$ 0.1} & 62.4 $\pm$ 0.2 \\
    ImageNet-1K (0.03 -- 0.97)   & \textbf{80.8 $\pm$ 0.1} & 74.2 $\pm$ 0.1 \\
    ImageNet-1K (0.1 -- 0.9)     & \textbf{84.3 $\pm$ 0.1} & 83.1 $\pm$ 0.1 \\
    ImageNet-1K (0.33 -- 0.67)   & \textbf{88.1 $\pm$ 0.1} & 87.8 $\pm$ 0.1 \\
    ImageNet-1K (0.5 -- 0.5)     & \textbf{88.5 $\pm$ 0.1} & 88.0 $\pm$ 0.1 \\
    CREMA-D (A-V-1-frame)        & \textbf{75.3 $\pm$ 0.9} & 70.1 $\pm$ 2.0 \\
    CREMA-D (A-V-4-frames)       & \textbf{80.0 $\pm$ 1.0} & 73.5 $\pm$ 1.8 \\
\bottomrule
\end{tabular}
\caption{Deep ensembles outperform hybrid approaches that use a late-fusion network as the multimodal component, on both synthetic and real-world datasets.}
\label{table:full_de_vs_i2m2}
\end{table}

For early-fusion concatenation, all input streams are merged at the pixel level before being passed to a single shared encoder. 
For homogeneous inputs (e.g., two RGB images), any spatial size mismatch is resolved by bilinearly resizing to the spatial dimensions of the first input, after which the streams are concatenated along the channel dimension and fed to a standard ResNet with the correspondingly widened first convolutional layer. 
For heterogeneous video–audio inputs, the single-channel log-mel spectrogram is bilinearly resized to the spatial dimensions of the video frames (H × W) and then tiled along the temporal axis to produce a (1, T, H, W) tensor, which is channel-concatenated with the RGB video tensor (3, T, H, W) to form a (4, T, H, W) input processed by a single 3D ResNet. 
In both cases, the resulting feature vector is passed to the same shared adaptor used by the other fusion variants.

The results for the early-fusion concatenation are provided in Table~\ref{table:full_de_vs_i2m2_early_fusion}. 
The multimodal network requires some additional parameters to project the 6-channel input to a feature map, but this adds negligibly more parameters compared to a unimodal baseline. 
Therefore, we compare I2M2 to a heterogeneous ensemble of size three in this scenario.
\begin{table}[htbp]
\centering
\begin{tabular}{l|l|l}
\toprule
    Dataset & HeDE(3) & I2M2 \\
    \hline
    CIFAR-100 (0.05 -- 0.95) & \textbf{81.0 $\pm$ 0.2} & 74.8 $\pm$ 0.3 \\
    CIFAR-100 (0.1 -- 0.9)   & \textbf{83.8 $\pm$ 0.2} & 79.9 $\pm$ 0.3 \\
    CIFAR-100 (0.2 -- 0.8)   & \textbf{86.2 $\pm$ 0.3} & 84.6 $\pm$ 0.3 \\
    CIFAR-100 (0.3 -- 0.7)   & \textbf{86.5 $\pm$ 0.2} & 86.4 $\pm$ 0.3 \\
    CIFAR-100 (0.5 -- 0.5)   & 85.9 $\pm$ 0.2 & \textbf{88.0 $\pm$ 0.2} \\
    CREMA-D (A-V-1-frame)     & \textbf{74.4 $\pm$ 1.3} & 70.2 $\pm$ 1.9 \\
    CREMA-D (A-V-4-frames)    & \textbf{79.5 $\pm$ 1.1} & 74.7 $\pm$ 1.5 \\
\bottomrule
\end{tabular}
\caption{At equal parameter count, deep ensembles outperform hybrid approaches that use an early-fusion network as the multimodal component, on both synthetic and real-world datasets.}
\label{table:full_de_vs_i2m2_early_fusion}
\end{table}
HeDEs outperform I2M2 in most settings, with the largest gap for imbalanced modalities. 
In the balanced CIFAR-100 (0.5–0.5) early-fusion case, I2M2 outperforms HeDE(3); however, scaling to HeDE(6) vs. two I2M2 models restores the advantage (89.6 $\pm$ 0.1 vs. 89.2 $\pm$ 0.2).

%% file: content/section_appendix_c_logits_vs_probs.tex
As mentioned in Sec.~\ref{sec:deep_ens_and_late_fusion}, logit
averaging consistently outperforms probability averaging in
heterogeneous deep ensembles. By default, we average the logits of
unimodal models directly without rescaling. However, one can first
calibrate each unimodal model independently by finding per-modality
temperatures $t_a$ and $t_v$ that minimize the validation loss via
grid search, and then average the resulting calibrated probabilities.

\begin{figure}[htbp]
\centering
\begin{subfigure}[t]{0.48\linewidth}
    \centering
    \includegraphics[width=\linewidth]{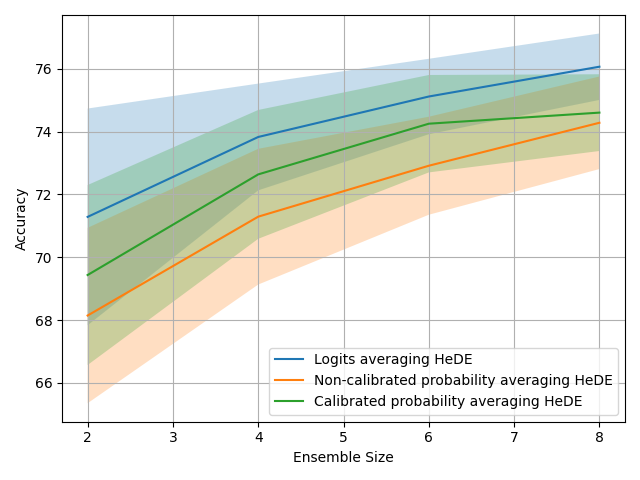}
    \caption{
    Logits-averaging ensembles yield strong classification results across all ensemble sizes.
    In contrast, probability-averaging ensembles with non-calibrated logits lead to worse performance.
    Probability-averaging ensembles with calibrated logits yield results between these two cases.
    }
    \label{fig:logits_vs_probs_vs_calibrated_diff_sizes}
\end{subfigure}
\hfill
\begin{subfigure}[t]{0.48\linewidth}
    \centering
    \includegraphics[width=\linewidth]{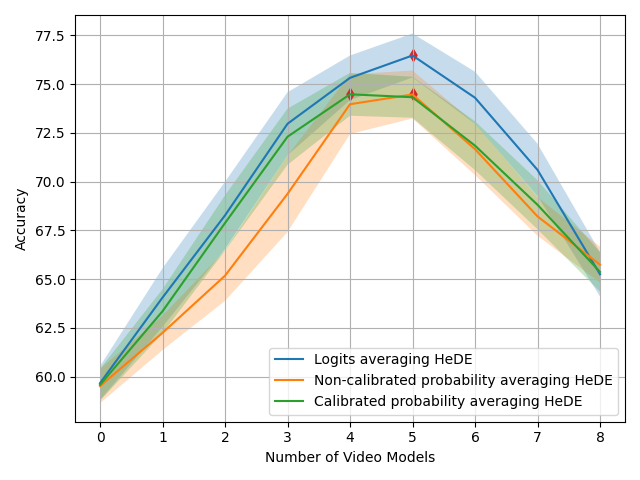}
    \caption{Accuracies for logits-averaging, calibrated and non-calibrated probability-averaging heterogeneous ensembles of size 8 with different unimodal models allocations. Findings repeat those from Fig.~\ref{fig:logits_vs_probs_vs_calibrated_diff_sizes}.}
    \label{fig:logits_vs_probs_vs_calibrated_ens_size_8}
\end{subfigure}
\caption{Performance of logits-averaging, calibrated and non-calibrated probability-averaging ensembles.}
\label{fig:crema_d_ens_size_2_8_calibration}
\end{figure}

Fig.~\ref{fig:logits_vs_probs_vs_calibrated_diff_sizes} compares
the three strategies on CREMA-D across ensemble sizes 2 to 8.
Logit averaging achieves the highest accuracy at every ensemble
size. Probability averaging without calibration performs worst,
and calibrated probability averaging falls between the two.
Fig.~\ref{fig:logits_vs_probs_vs_calibrated_ens_size_8} shows the
same comparison for a heterogeneous ensemble of size 8 as the number
of stronger unimodal models varies; the ranking of the three
strategies is consistent across all allocations.

%% file: content/section_appendix_calibration.tex
One outcome of applying logit-averaging heterogeneous deep ensembles
is that they yield well-calibrated predictions. We assess this by
evaluating the Expected Calibration Error (ECE)~\cite{naeini_2015_ece}
and plotting reliability diagrams~\cite{guo_2017_calibration_of_nns}
for models trained on synthetic ImageNet-1K and CREMA-D datasets
with varying levels of modality imbalance. The ECE summarises
calibration as a weighted average of per-bin deviations from
perfect calibration:
\begin{gather*}
\mathrm{ECE} = \sum_b \frac{|B_b|}{n}
  \left| \mathrm{acc}(B_b) - \mathrm{conf}(B_b) \right|,
\end{gather*}
where $B_b$ is the set of test samples in bin $b$ and $n$ is the
total sample count. A model is perfectly calibrated when
$\mathrm{ECE} = 0$.

\begin{figure}[htbp]
    \centering
    \includegraphics[width=\linewidth]{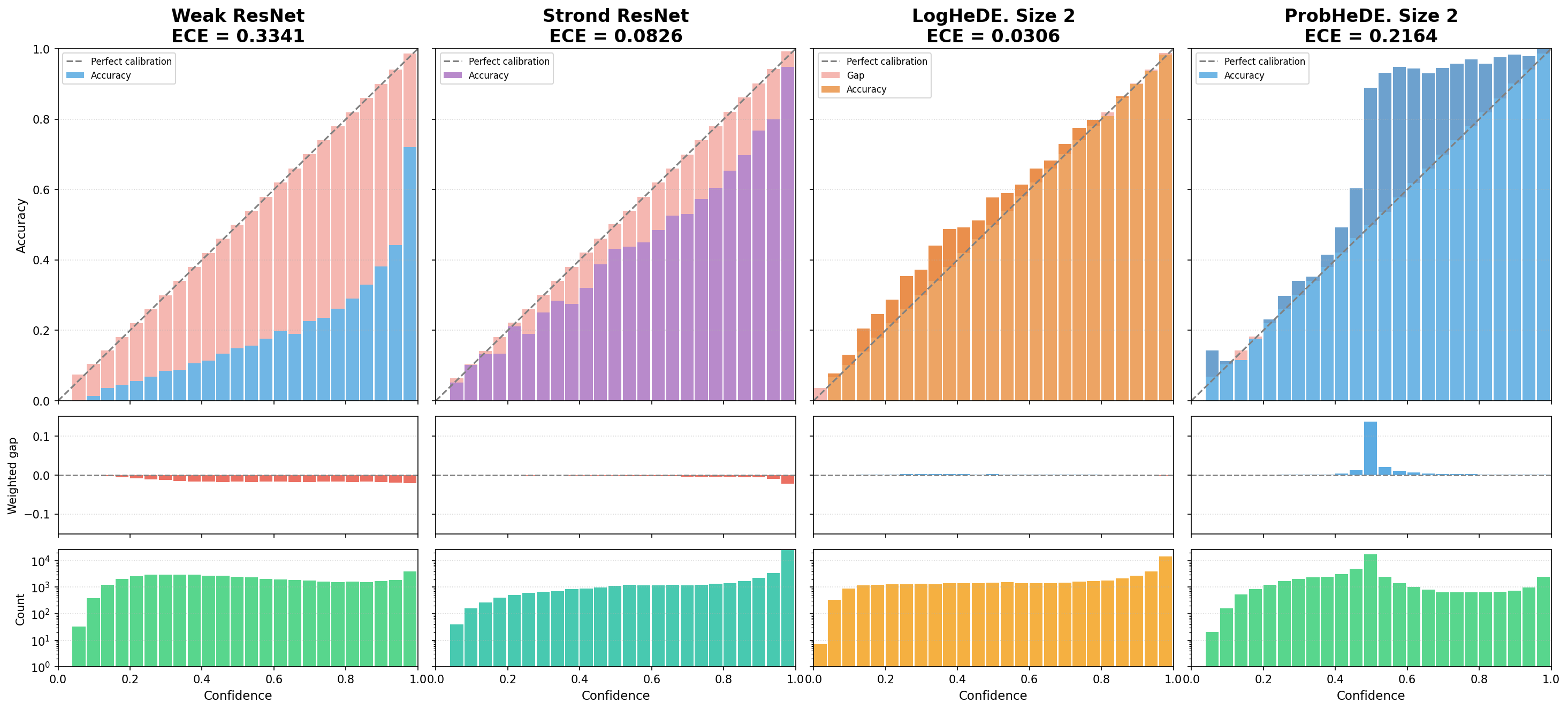}
    \caption{Reliability diagrams for ImageNet-1K (0.03\,--\,0.97):
    weaker unimodal model (left), stronger unimodal model
    (centre-left), LogHeDE (centre-right), and ProbHeDE (right).}
    \label{fig:ece_in_97_03_non_calib_ens_size_2}
\end{figure}

\begin{figure}[htbp]
    \centering
    \includegraphics[width=\linewidth]{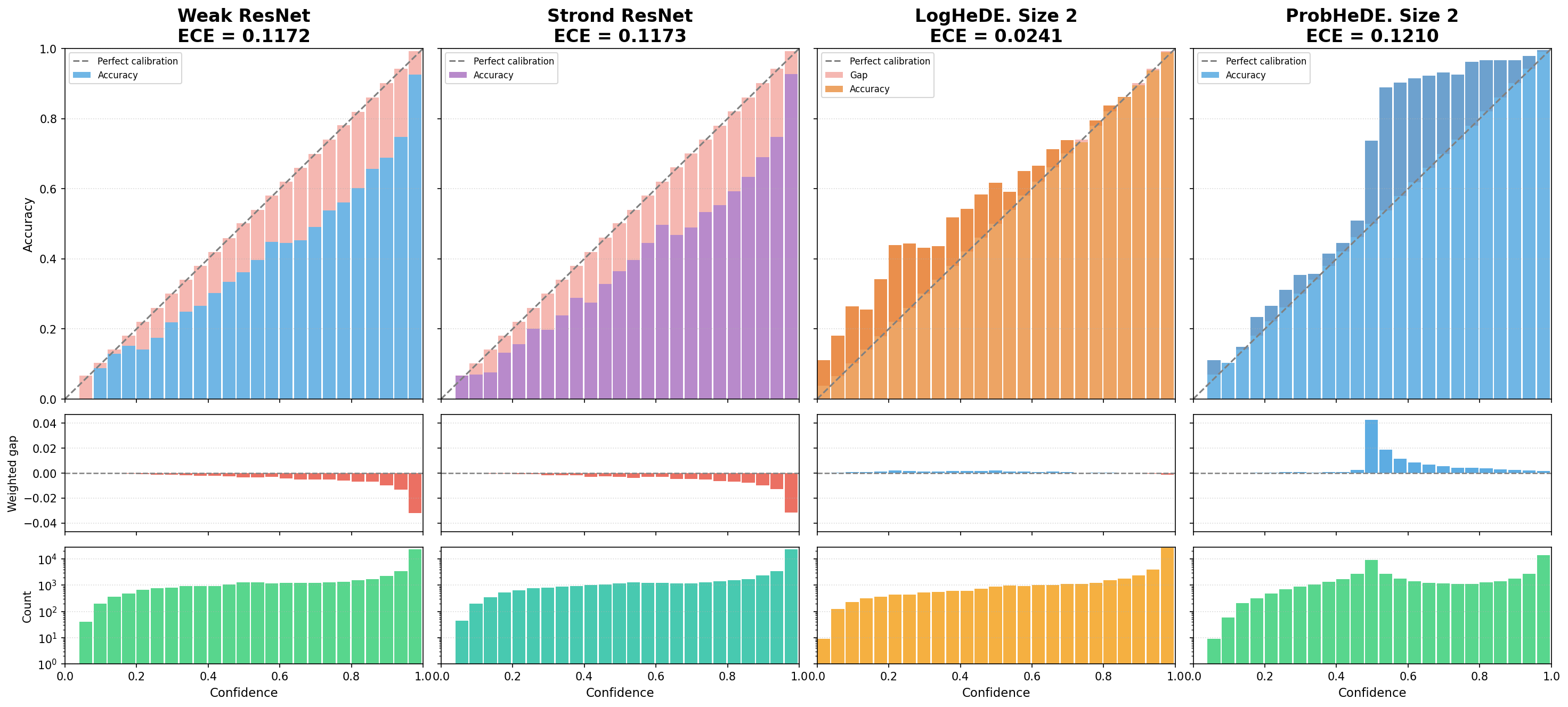}
    \caption{Reliability diagrams for ImageNet-1K (0.50\,--\,0.50):
    layout as in Fig.~\ref{fig:ece_in_97_03_non_calib_ens_size_2}.}
    \label{fig:ece_in_05_05_non_calib_ens_size_2}
\end{figure}

\begin{figure}[htbp]
    \centering
    \includegraphics[width=\linewidth]{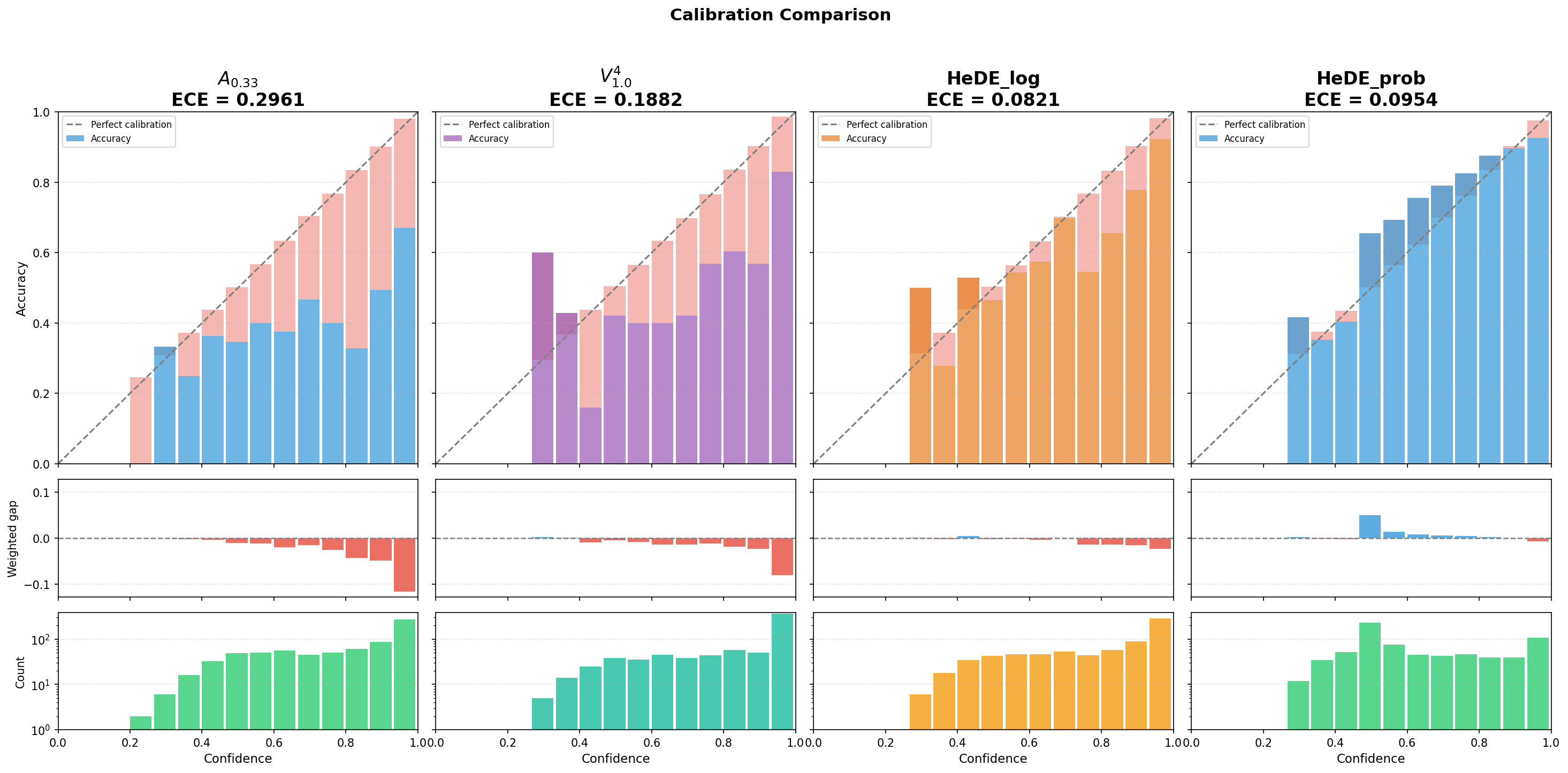}
    \caption{Reliability diagrams for CREMA-D
    (A$_{0.33}$ / V$_{1.0}^{4}$):
    layout as in Fig.~\ref{fig:ece_in_97_03_non_calib_ens_size_2}.}
    \label{fig:ece_crema_a_0.33_v_1.0_f_4_ens_size_2}
\end{figure}

\begin{figure}[htbp]
    \centering
    \includegraphics[width=\linewidth]{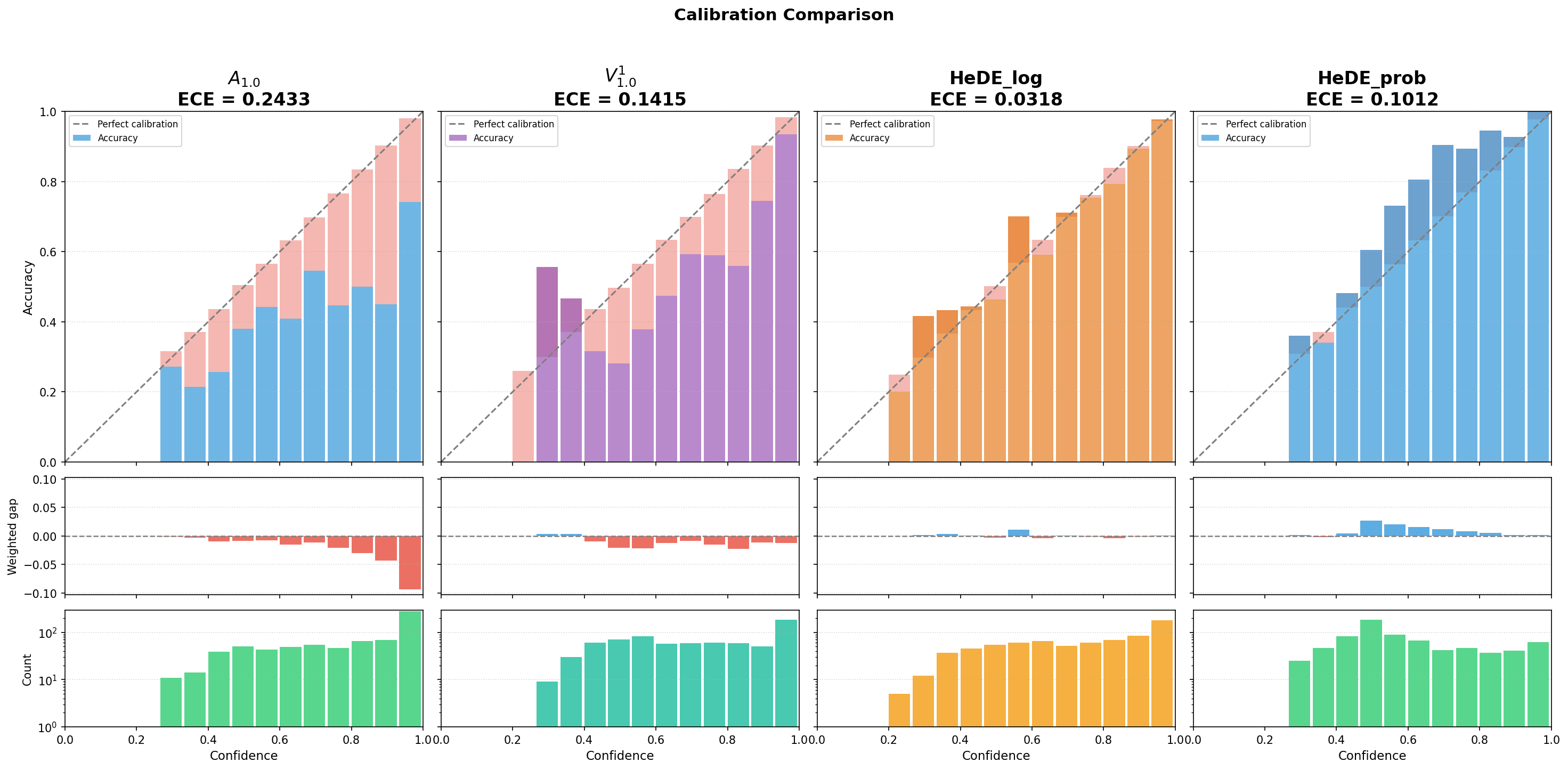}
    \caption{Reliability diagrams for CREMA-D
    (A$_{1.0}$ / V$_{1.0}^{1}$):
    layout as in Fig.~\ref{fig:ece_in_97_03_non_calib_ens_size_2}.}
    \label{fig:ece_crema_a_1.0_v_1.0_f_1_ens_size_2}
\end{figure}

Each panel in
Figs.~\ref{fig:ece_in_97_03_non_calib_ens_size_2}--\ref{fig:ece_crema_a_1.0_v_1.0_f_1_ens_size_2}
comprises three vertically stacked sub-panels: a reliability diagram
(top), in which bars show the observed accuracy per confidence bin
and the dashed diagonal represents perfect calibration; a signed
calibration-gap plot (middle), in which each bar shows the
ECE-weighted difference $({|B_b|}/{n})\cdot(\mathrm{acc}(B_b) -
\mathrm{conf}(B_b))$, with blue indicating underconfidence and red
indicating overconfidence; and a log-scale confidence histogram
(bottom). All panels within a figure share the same histogram
y-axis scale to enable direct comparison across models. The ECE
is reported in each panel title.

Table~\ref{tab:ece} summarises the ECE across all configurations.
On ImageNet-1K, logits-averaing heterogeneous deep ensembles consistently achieves the lowest ECE, substantially improving over both unimodal models regardless of imbalance level or ensemble size. 
On CREMA-D, the pattern depends on which modality is weaker: when the audio modality is weaker (first four CREMA-D rows), LogHeDE produces predictions that are better calibrated or on par with ProbHeDE in most configurations; when the video modality is weaker (last two rows), ProbHeDE achieves lower ECE. In both cases, at least one fusion strategy improves calibration over the
individual unimodal models.

\begin{table}[t]
\centering
\setlength{\tabcolsep}{5pt}
\begin{tabular}{l|c|c|c|c|c|c}
\toprule
Dataset & UM$_1$ / UM$_2$ & Log(2) & Prob(2)
        & Log(10) & Prob(10) \\
\midrule
IN-1K (0.03 -- 0.97)             & 0.334 / 0.083 & \textbf{0.031} & 0.216          & \textbf{0.028} & 0.127 \\
IN-1K (0.50 -- 0.50)             & 0.117 / 0.117 & \textbf{0.024} & 0.121          & \textbf{0.029} & 0.172 \\
\midrule
CREMA-D (A$_{0.33}$ / V$_{1.0}^{1}$)  & 0.296 / 0.141 & 0.091          & \textbf{0.077} & \textbf{0.054} & 0.085 \\
CREMA-D (A$_{1.0}$  / V$_{1.0}^{1}$)  & 0.243 / 0.141 & \textbf{0.032} & 0.101          & \textbf{0.037} & 0.105 \\
CREMA-D (A$_{0.33}$ / V$_{1.0}^{4}$)  & 0.296 / 0.188 & \textbf{0.082} & 0.095          & 0.092          & \textbf{0.081} \\
CREMA-D (A$_{1.0}$  / V$_{1.0}^{4}$)  & 0.243 / 0.188 & \textbf{0.047} & 0.115          & \textbf{0.066} & 0.102 \\
CREMA-D (A$_{1.0}$  / V$_{0.1}^{1}$)  & 0.243 / 0.344 & 0.127          & \textbf{0.080} & 0.142          & \textbf{0.072} \\
CREMA-D (A$_{1.0}$  / V$_{0.33}^{1}$) & 0.243 / 0.474 & 0.260          & \textbf{0.128} & 0.193          & \textbf{0.078} \\
\bottomrule
\end{tabular}
\caption{Expected Calibration Error for individual
unimodal models and two HeDE fusion strategies, without post-hoc
calibration. 
\texttt{Log($K$)} and \texttt{Prob($K$)} denote logit-averaging and probability-averaging heterogeneous deep ensemble of size $K$.
\texttt{UM}$_i$: ECE of the unimodal model trained on
modality $i$. The number in parentheses denotes ensemble size $K$;
$K=10$ uses loss-proportional copies of each modality's logits.
For CREMA-D, \texttt{A} and \texttt{V} denote the audio and video
modality, respectively; subscripts indicate the fraction of training
data used, and superscripts on \texttt{V} indicate the number of
input frames.}
\label{tab:ece}
\end{table}

%% file: content/section_appendix_d_scaling_laws.tex
Beyond predicting the optimal proportion, we investigate whether the
asymptotic ensemble performance can be estimated without training any
ensembles. Across all 24 bimodal configurations described in
Sec.~\ref{sec:scaling_laws}, we fit a linear regression predicting
$\min_{\theta} L_{\infty}(\theta)$ from the logarithmic unimodal
losses of the two component modalities. The resulting model achieves
$R^2 = 0.96$, indicating that the asymptotic performance of
optimally allocated heterogeneous ensembles is largely determined by
the individual unimodal losses. This complements the heuristic from
Sec.~\ref{subsec:number_of_networks}, which predicts the optimal
allocation: together, the two results suggest that unimodal
validation losses are sufficient to estimate both the composition and
the expected performance of heterogeneous deep ensembles without
training any.

\begin{figure}[htbp]
\centering
\begin{subfigure}[t]{0.48\linewidth}
    \centering
    \includegraphics[width=\linewidth]{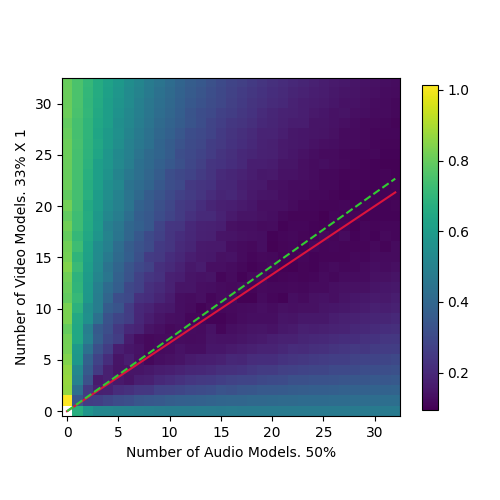}
    \caption{Test log-loss for HeDEs. Red line: optimal proportion;
    green dashed: heuristic prediction.}
    \label{fig:cream_log_loss_vs_ens_size_a_05_v_nf_1_df_033}
\end{subfigure}
\hfill
\begin{subfigure}[t]{0.4\linewidth}
    \centering
    \includegraphics[width=\linewidth]{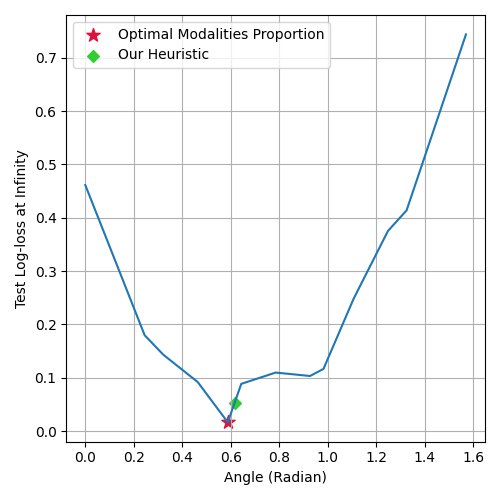}
    \caption{Asymptotic log-loss $L_{\infty}(\theta)$ vs.\ angle
    $\theta$. Red star: optimal; green diamond: heuristic.}
    \label{fig:loss_vs_angle_a_05_v_nf_1_df_033}
\end{subfigure}
\caption{Scaling laws on CREMA-D. Audio model trained on 50\% of
the dataset; video model trained on 33\% with 1 input frame.}
\label{fig:scaling_laws_a_05_v_033}
\end{figure}

\begin{figure}[htbp]
\centering
\begin{subfigure}[t]{0.48\linewidth}
    \centering
    \includegraphics[width=\linewidth]{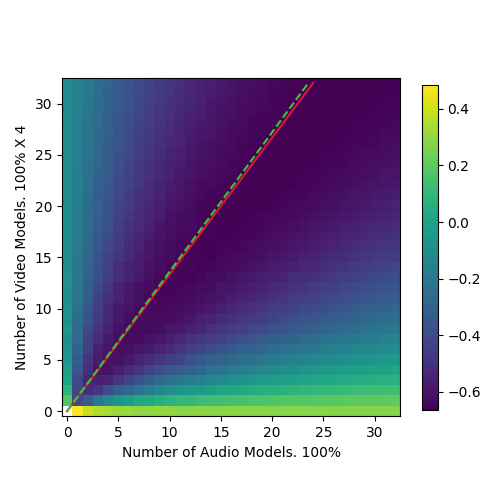}
    \caption{Test log-loss for HeDEs. Red line: optimal proportion;
    green dashed: heuristic prediction.}
    \label{fig:cream_log_loss_vs_ens_size_a_1_v_nf_4_df_1}
\end{subfigure}
\hfill
\begin{subfigure}[t]{0.4\linewidth}
    \centering
    \includegraphics[width=\linewidth]{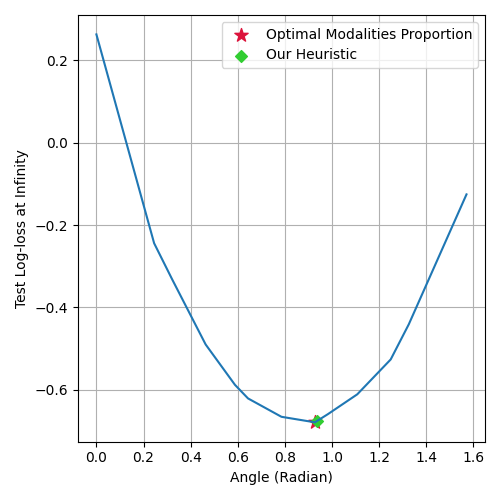}
    \caption{Asymptotic log-loss $L_{\infty}(\theta)$ vs.\ angle
    $\theta$. Red star: optimal; green diamond: heuristic.}
    \label{fig:loss_vs_angle_a_1_v_nf_4_df_1}
\end{subfigure}
\caption{Scaling laws on CREMA-D. Both models trained on the full
dataset; video model operates on 4 input frames.}
\label{fig:scaling_laws_a_1_v_1}
\end{figure}

To demonstrate that the scaling behaviour observed in the main text
generalises across imbalance levels, we present analogous figures for
two additional modality configurations.
Fig.~\ref{fig:scaling_laws_a_05_v_033} shows the case where the
audio model is trained on 50\% of the dataset and the video model on
33\%; Fig.~\ref{fig:scaling_laws_a_1_v_1} shows the case where both
models are trained on the full dataset with the video model operating
on 4 frames. In both cases, the heuristic closely tracks the
brute-force-optimal proportion.

\begin{figure}[htbp]
\centering
\begin{subfigure}[t]{0.32\linewidth}
    \centering
    \includegraphics[width=\linewidth]{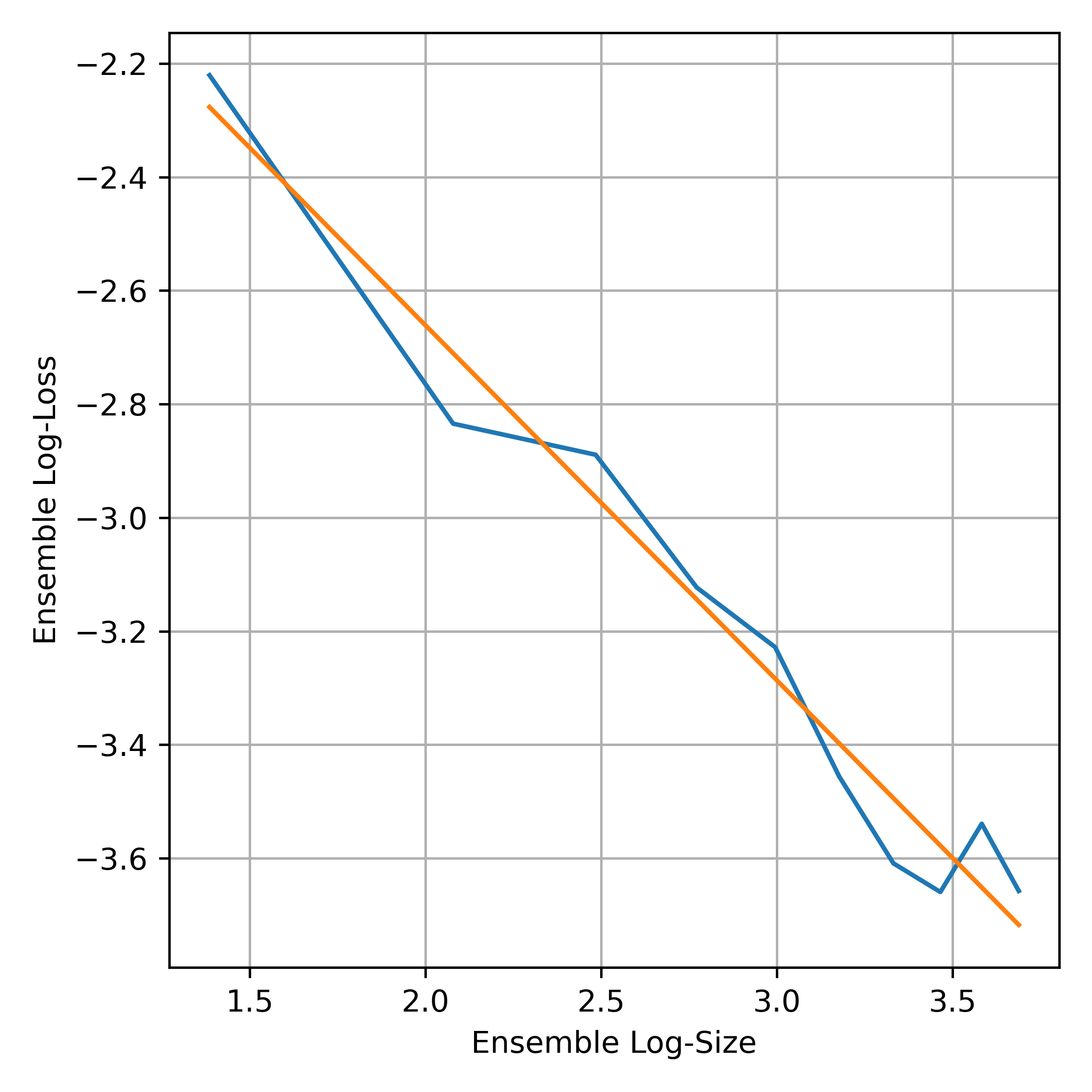}
    \caption{$\theta = -0.322$}
    \label{fig:pl_angle_0322}
\end{subfigure}
\hfill
\begin{subfigure}[t]{0.32\linewidth}
    \centering
    \includegraphics[width=\linewidth]{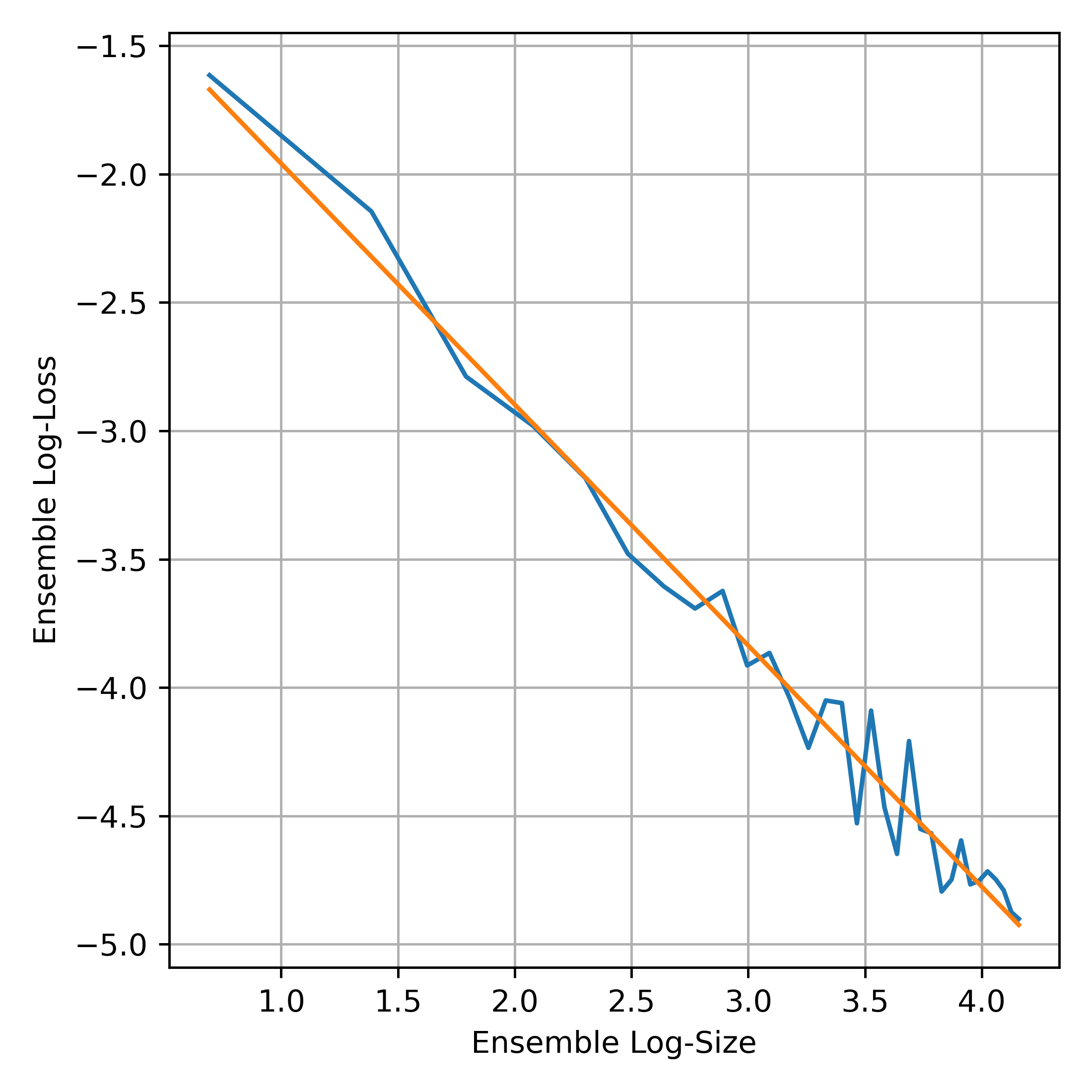}
    \caption{$\theta = -0.785$}
    \label{fig:pl_angle_0785}
\end{subfigure}
\hfill
\begin{subfigure}[t]{0.32\linewidth}
    \centering
    \includegraphics[width=\linewidth]{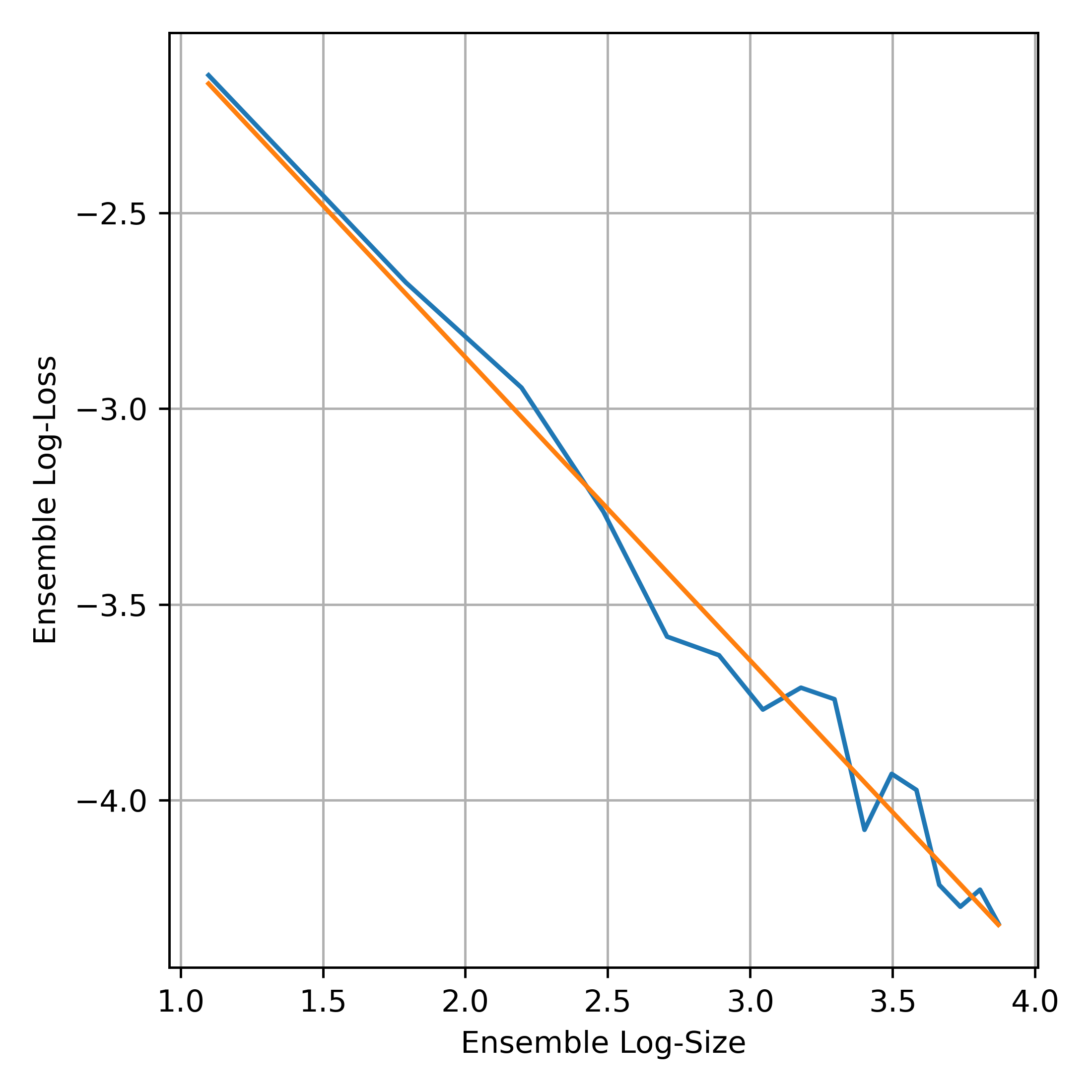}
    \caption{$\theta = -1.107$}
    \label{fig:pl_angle_1107}
\end{subfigure}
\caption{One-dimensional power law fits at three representative
angles for the same modality pair as in the main text.}
\label{fig:one_dim_pl_angles}
\end{figure}

Finally, Fig.~\ref{fig:one_dim_pl_angles} shows examples of
one-dimensional power law fits along individual rays (fixed
proportions) for the same modality pair as in the main text (audio
and video models trained on the full dataset, video with 1 input
frame). The fits confirm that the power law form
$L_N(\theta) = L_{\infty}(\theta) + A(\theta)/N^{\alpha(\theta)}$
holds well at each angle individually.

%% file: content/section_appendix_f_cost_comparison.tex
%
%
All experiments were conducted on a SLURM-managed cluster equipped with NVIDIA H100 GPUs, with 14 CPU cores allocated per GPU. Training times and peak GPU memory reported in Table~\ref{tab:training_times} were obtained using the \texttt{sacct} command.

A heterogeneous deep ensemble of size $N_\text{ens}$ consists of
$N_\text{ens}$ unimodal networks, each with its own encoder and
classification head. A bimodal late-fusion network consists of two
encoders and a joint classifier operating on the concatenated
features. For $N_\text{ens} = 2$, the total parameter count of a
HeDE is therefore equal to that of a single late-fusion
network.
The same equivalence holds for all imbalance-aware late-fusion
baselines we compare against (OGM, MMPareto, AUG, InfoReg,
ReconBoost), as these methods modify the training procedure but do
not add parameters to the base late-fusion architecture, with the
exception of AUG which introduces a small additional boosting modules.
Intermediate-fusion methods do add parameters:
Table~\ref{tab:fusion_overhead} reports the overhead introduced by
MMTM blocks and cross-attention blocks for the architectures used in
our experiments.

\begin{table}[htbp]
\centering
\setlength{\tabcolsep}{5pt}
\begin{tabular}{l|l|r}
\toprule
Architecture & Fusion blocks & Overhead \\
\midrule
\multicolumn{3}{l}{\textit{MMTM}} \\
\quad ResNet-18  & 1 (layer 4)        & 1.05\,M \\
\quad ResNet-18  & 2 (layers 3--4)    & 1.31\,M \\
\quad ResNet-50  & 1 (layer 4)        & 16.78\,M \\
\quad ResNet-50  & 2 (layers 3--4)    & 20.98\,M \\
\midrule
\multicolumn{3}{l}{\textit{Cross-attention}} \\
\quad ViT-Tiny   & 1 block & 0.30\,M \\
\quad ViT-Small  & 1 block & 1.18\,M \\
\quad ViT-Base   & 1 block & 4.73\,M \\
\quad ViT-Large  & 1 block & 8.40\,M \\
\bottomrule
\end{tabular}
\caption{Additional parameters introduced by intermediate-fusion
blocks. For MMTM, the overhead depends on the insertion layers; for
cross-attention, it scales with the embedding dimension and is
identical across all insertion points.}
\label{tab:fusion_overhead}
\end{table}

At inference time, the unimodal models in a HeDE can be evaluated
independently and their logits averaged, making the inference cost
comparable to that of a late-fusion network of equal parameter count.

Table~\ref{tab:training_times} reports wall-clock training times and
peak RAM memory for all methods. Training a HeDE requires training
each unimodal component separately; the total cost is the sum of the
individual unimodal training times. On CREMA-D, a HeDE(2) comprising
one audio and one video model trains in approximately 1h\,14m, which
is comparable to or faster than most imbalance-aware baselines. On
CIFAR-100 and ImageNet, where both modalities are derived from the
same source (0.5--0.5 split), a HeDE(2) costs approximately twice
the unimodal training time. Since unimodal models are independent,
their training can also be fully parallelised across GPUs.

\begin{table}[htbp]
\centering
\setlength{\tabcolsep}{4pt}
\begin{tabular}{l|r|r|r|r|r|r}
\toprule
& \multicolumn{2}{c|}{CREMA-D}
& \multicolumn{2}{c|}{CIFAR-100}
& \multicolumn{2}{c}{ImageNet} \\
\cmidrule(lr){2-3} \cmidrule(lr){4-5} \cmidrule(lr){6-7}
Method & Time & Mem & Time & Mem & Time & Mem \\
\midrule
Late-fusion     & 0:17  & 44\,Gb & 0:09    & 11\,Gb     & 20:24    & 72\,Gb \\
OGM             & 0:31  & 32\,Gb & 0:10  & 20\,Gb & 1d\,3:03 & 43\,Gb \\
MMPareto        & 0:30  & 26\,Gb & 0:13  & 21\,Gb & 2d\,9:46 & 44\,Gb \\
AUG             & 0:54  & 48\,Gb & 0:11  & 28\,Gb & 1d\,2:37 & 60\,Gb \\
InfoReg         & 0:37  & 24\,Gb & 0:19  & 22\,Gb & 3d\,8:28 & 46\,Gb \\
ReconBoost      & 4:00  & 22\,Gb & 0:07  & 19\,Gb & 2d\,4:11 & 30\,Gb \\
LFM             & 4:12  & 31\,Gb & --    & --     & --       & -- \\
\midrule
MMTM-1          & 0:19    & 44\,Gb     & 0:09    & 11\,Gb     & 20:35    & 74\,Gb \\
MMTM-2          & 0:17    & 46\,Gb     & 0:09    & 11\,Gb     & 20:38    & 77\,Gb \\
ViT + CA (4 GPUs) & --    & --     & --    & --     & 19:30    & 95\,Gb \\
\midrule
Unimodal audio  & 0:22  & 17\,Gb & --    & --     & --       & -- \\
Unimodal video  & 0:52  & 50\,Gb & 0:08  & 49\,Gb & 12:42    & 56\,Gb \\
HeDE(2)         &$\sim$1:14& --  &$\sim$0:16& --  &$\sim$25:23& -- \\
\bottomrule
\end{tabular}
\caption{Wall-clock training time (h:mm or d\,h:mm) and peak RAM
memory (Mem) on a single H100 GPU unless otherwise noted.
HeDE(2) time is the sum of its
unimodal components. For CIFAR-100 and ImageNet, both modalities
are derived from the same source (0.5--0.5 split), so each component
has equal training cost. Kinetics-400 late-fusion training
(8$\times$\,H100, not shown) takes approximately 1d\,10h;
trimodal ImageNet late-fusion (1$\times$\,H100) takes approximately
1d\,1h.}
\label{tab:training_times}
\end{table}

%% file: checklist.tex
\section*{NeurIPS Paper Checklist}

\begin{enumerate}

\item {\bf Claims}
    \item[] Question: Do the main claims made in the abstract and introduction accurately reflect the paper's contributions and scope?
    \item[] Answer: \answerYes{} 
    \item[] Justification: The abstract states four contributions, each supported by experiments in Sections~\ref{sec:results} and \ref{sec:scaling_laws} and Appendices~\ref{sec:appendix_additional_results} and ~\ref{sec:appendix_scaling_laws}. Limitations are discussed in Section~\ref{sec:discussion}.
    \item[] Guidelines:
    \begin{itemize}
        \item The answer \answerNA{} means that the abstract and introduction do not include the claims made in the paper.
        \item The abstract and/or introduction should clearly state the claims made, including the contributions made in the paper and important assumptions and limitations. A \answerNo{} or \answerNA{} answer to this question will not be perceived well by the reviewers. 
        \item The claims made should match theoretical and experimental results, and reflect how much the results can be expected to generalize to other settings. 
        \item It is fine to include aspirational goals as motivation as long as it is clear that these goals are not attained by the paper. 
    \end{itemize}

\item {\bf Limitations}
    \item[] Question: Does the paper discuss the limitations of the work performed by the authors?
    \item[] Answer: \answerYes{} 
    \item[] Justification: Section~\ref{sec:discussion} discusses the limitations.
    \item[] Guidelines:
    \begin{itemize}
        \item The answer \answerNA{} means that the paper has no limitation while the answer \answerNo{} means that the paper has limitations, but those are not discussed in the paper. 
        \item The authors are encouraged to create a separate ``Limitations'' section in their paper.
        \item The paper should point out any strong assumptions and how robust the results are to violations of these assumptions (e.g., independence assumptions, noiseless settings, model well-specification, asymptotic approximations only holding locally). The authors should reflect on how these assumptions might be violated in practice and what the implications would be.
        \item The authors should reflect on the scope of the claims made, e.g., if the approach was only tested on a few datasets or with a few runs. In general, empirical results often depend on implicit assumptions, which should be articulated.
        \item The authors should reflect on the factors that influence the performance of the approach. For example, a facial recognition algorithm may perform poorly when image resolution is low or images are taken in low lighting. Or a speech-to-text system might not be used reliably to provide closed captions for online lectures because it fails to handle technical jargon.
        \item The authors should discuss the computational efficiency of the proposed algorithms and how they scale with dataset size.
        \item If applicable, the authors should discuss possible limitations of their approach to address problems of privacy and fairness.
        \item While the authors might fear that complete honesty about limitations might be used by reviewers as grounds for rejection, a worse outcome might be that reviewers discover limitations that aren't acknowledged in the paper. The authors should use their best judgment and recognize that individual actions in favor of transparency play an important role in developing norms that preserve the integrity of the community. Reviewers will be specifically instructed to not penalize honesty concerning limitations.
    \end{itemize}

\item {\bf Theory assumptions and proofs}
    \item[] Question: For each theoretical result, does the paper provide the full set of assumptions and a complete (and correct) proof?
    \item[] Answer: \answerNA{} 
    \item[] Justification: This paper does not include formal theoretical results.
    \item[] Guidelines:
    \begin{itemize}
        \item The answer \answerNA{} means that the paper does not include theoretical results. 
        \item All the theorems, formulas, and proofs in the paper should be numbered and cross-referenced.
        \item All assumptions should be clearly stated or referenced in the statement of any theorems.
        \item The proofs can either appear in the main paper or the supplemental material, but if they appear in the supplemental material, the authors are encouraged to provide a short proof sketch to provide intuition. 
        \item Inversely, any informal proof provided in the core of the paper should be complemented by formal proofs provided in appendix or supplemental material.
        \item Theorems and Lemmas that the proof relies upon should be properly referenced. 
    \end{itemize}

    \item {\bf Experimental result reproducibility}
    \item[] Question: Does the paper fully disclose all the information needed to reproduce the main experimental results of the paper to the extent that it affects the main claims and/or conclusions of the paper (regardless of whether the code and data are provided or not)?
    \item[] Answer: \answerYes{} 
    \item[] Justification: All training details, hyperparameters, and data splits are provided in Appendix~\ref{sec:appendix_training_details}. Baseline results were reproduced using the authors' original code repositories. Modifications to baseline implementations are documented in Appendix~\ref{sec:appendix_training_details}.
    \item[] Guidelines:
    \begin{itemize}
        \item The answer \answerNA{} means that the paper does not include experiments.
        \item If the paper includes experiments, a \answerNo{} answer to this question will not be perceived well by the reviewers: Making the paper reproducible is important, regardless of whether the code and data are provided or not.
        \item If the contribution is a dataset and\slash or model, the authors should describe the steps taken to make their results reproducible or verifiable. 
        \item Depending on the contribution, reproducibility can be accomplished in various ways. For example, if the contribution is a novel architecture, describing the architecture fully might suffice, or if the contribution is a specific model and empirical evaluation, it may be necessary to either make it possible for others to replicate the model with the same dataset, or provide access to the model. In general. releasing code and data is often one good way to accomplish this, but reproducibility can also be provided via detailed instructions for how to replicate the results, access to a hosted model (e.g., in the case of a large language model), releasing of a model checkpoint, or other means that are appropriate to the research performed.
        \item While NeurIPS does not require releasing code, the conference does require all submissions to provide some reasonable avenue for reproducibility, which may depend on the nature of the contribution. For example
        \begin{enumerate}
            \item If the contribution is primarily a new algorithm, the paper should make it clear how to reproduce that algorithm.
            \item If the contribution is primarily a new model architecture, the paper should describe the architecture clearly and fully.
            \item If the contribution is a new model (e.g., a large language model), then there should either be a way to access this model for reproducing the results or a way to reproduce the model (e.g., with an open-source dataset or instructions for how to construct the dataset).
            \item We recognize that reproducibility may be tricky in some cases, in which case authors are welcome to describe the particular way they provide for reproducibility. In the case of closed-source models, it may be that access to the model is limited in some way (e.g., to registered users), but it should be possible for other researchers to have some path to reproducing or verifying the results.
        \end{enumerate}
    \end{itemize}

\item {\bf Open access to data and code}
    \item[] Question: Does the paper provide open access to the data and code, with sufficient instructions to faithfully reproduce the main experimental results, as described in supplemental material?
    \item[] Answer: \answerYes{} 
    \item[] Justification: Code to reproduce all experiments is provided as supplementary material. All datasets used (CIFAR-100, ImageNet-1K, CREMA-D, Sarcasm, Kinetics-400) are publicly available.
    \item[] Guidelines:
    \begin{itemize}
        \item The answer \answerNA{} means that paper does not include experiments requiring code.
        \item Please see the NeurIPS code and data submission guidelines (\url{https://neurips.cc/public/guides/CodeSubmissionPolicy}) for more details.
        \item While we encourage the release of code and data, we understand that this might not be possible, so \answerNo{} is an acceptable answer. Papers cannot be rejected simply for not including code, unless this is central to the contribution (e.g., for a new open-source benchmark).
        \item The instructions should contain the exact command and environment needed to run to reproduce the results. See the NeurIPS code and data submission guidelines (\url{https://neurips.cc/public/guides/CodeSubmissionPolicy}) for more details.
        \item The authors should provide instructions on data access and preparation, including how to access the raw data, preprocessed data, intermediate data, and generated data, etc.
        \item The authors should provide scripts to reproduce all experimental results for the new proposed method and baselines. If only a subset of experiments are reproducible, they should state which ones are omitted from the script and why.
        \item At submission time, to preserve anonymity, the authors should release anonymized versions (if applicable).
        \item Providing as much information as possible in supplemental material (appended to the paper) is recommended, but including URLs to data and code is permitted.
    \end{itemize}

\item {\bf Experimental setting/details}
    \item[] Question: Does the paper specify all the training and test details (e.g., data splits, hyperparameters, how they were chosen, type of optimizer) necessary to understand the results?
    \item[] Answer: \answerYes{} 
    \item[] Justification: Full training details including optimizers, learning rate schedules, batch sizes, augmentation strategies, and data preprocessing are provided in Appendix~\ref{sec:appendix_training_details}. The synthetic dataset construction is described in Section~\ref{sec:synthetic_data}. Computational costs are reported in Appendix~\ref{sec:appendix_cost_comparisons}.
    \item[] Guidelines:
    \begin{itemize}
        \item The answer \answerNA{} means that the paper does not include experiments.
        \item The experimental setting should be presented in the core of the paper to a level of detail that is necessary to appreciate the results and make sense of them.
        \item The full details can be provided either with the code, in appendix, or as supplemental material.
    \end{itemize}

\item {\bf Experiment statistical significance}
    \item[] Question: Does the paper report error bars suitably and correctly defined or other appropriate information about the statistical significance of the experiments?
    \item[] Answer: \answerYes{} 
    \item[] Justification: Results are reported as mean $\pm$ standard deviation across multiple runs (5 runs for CREMA-D, 10 ensembles for the ensemble-size-2 analysis)
    \item[] Guidelines:
    \begin{itemize}
        \item The answer \answerNA{} means that the paper does not include experiments.
        \item The authors should answer \answerYes{} if the results are accompanied by error bars, confidence intervals, or statistical significance tests, at least for the experiments that support the main claims of the paper.
        \item The factors of variability that the error bars are capturing should be clearly stated (for example, train/test split, initialization, random drawing of some parameter, or overall run with given experimental conditions).
        \item The method for calculating the error bars should be explained (closed form formula, call to a library function, bootstrap, etc.)
        \item The assumptions made should be given (e.g., Normally distributed errors).
        \item It should be clear whether the error bar is the standard deviation or the standard error of the mean.
        \item It is OK to report 1-sigma error bars, but one should state it. The authors should preferably report a 2-sigma error bar than state that they have a 96\% CI, if the hypothesis of Normality of errors is not verified.
        \item For asymmetric distributions, the authors should be careful not to show in tables or figures symmetric error bars that would yield results that are out of range (e.g., negative error rates).
        \item If error bars are reported in tables or plots, the authors should explain in the text how they were calculated and reference the corresponding figures or tables in the text.
    \end{itemize}

\item {\bf Experiments compute resources}
    \item[] Question: For each experiment, does the paper provide sufficient information on the computer resources (type of compute workers, memory, time of execution) needed to reproduce the experiments?
    \item[] Answer: \answerYes{} 
    \item[] Justification: All experiments were conducted on H100 GPUs. Wall-clock training times and peak RAM memory for all methods and datasets are reported in Appendix~\ref{sec:appendix_cost_comparisons} (Table~\ref{tab:training_times}). Multi-GPU configurations are noted.
    \item[] Guidelines:
    \begin{itemize}
        \item The answer \answerNA{} means that the paper does not include experiments.
        \item The paper should indicate the type of compute workers CPU or GPU, internal cluster, or cloud provider, including relevant memory and storage.
        \item The paper should provide the amount of compute required for each of the individual experimental runs as well as estimate the total compute. 
        \item The paper should disclose whether the full research project required more compute than the experiments reported in the paper (e.g., preliminary or failed experiments that didn't make it into the paper). 
    \end{itemize}
    
\item {\bf Code of ethics}
    \item[] Question: Does the research conducted in the paper conform, in every respect, with the NeurIPS Code of Ethics \url{https://neurips.cc/public/EthicsGuidelines}?
    \item[] Answer: \answerYes{} 
    \item[] Justification: The research conforms with the NeurIPS Code of Ethics. The work is a methodological study on multimodal classification using publicly available datasets. Potential societal implications are discussed in Section~\ref{sec:discussion}.
    \item[] Guidelines:
    \begin{itemize}
        \item The answer \answerNA{} means that the authors have not reviewed the NeurIPS Code of Ethics.
        \item If the authors answer \answerNo, they should explain the special circumstances that require a deviation from the Code of Ethics.
        \item The authors should make sure to preserve anonymity (e.g., if there is a special consideration due to laws or regulations in their jurisdiction).
    \end{itemize}

\item {\bf Broader impacts}
    \item[] Question: Does the paper discuss both potential positive societal impacts and negative societal impacts of the work performed?
    \item[] Answer: \answerYes{} 
    \item[] Justification: Section~\ref{sec:discussion} discusses potential societal impacts, noting that while the work does not introduce direct societal risks, improved multimodal classifiers could be applied in sensitive domains such as surveillance or medical diagnosis, where careful deployment considerations apply.
    \item[] Guidelines:
    \begin{itemize}
        \item The answer \answerNA{} means that there is no societal impact of the work performed.
        \item If the authors answer \answerNA{} or \answerNo, they should explain why their work has no societal impact or why the paper does not address societal impact.
        \item Examples of negative societal impacts include potential malicious or unintended uses (e.g., disinformation, generating fake profiles, surveillance), fairness considerations (e.g., deployment of technologies that could make decisions that unfairly impact specific groups), privacy considerations, and security considerations.
        \item The conference expects that many papers will be foundational research and not tied to particular applications, let alone deployments. However, if there is a direct path to any negative applications, the authors should point it out. For example, it is legitimate to point out that an improvement in the quality of generative models could be used to generate Deepfakes for disinformation. On the other hand, it is not needed to point out that a generic algorithm for optimizing neural networks could enable people to train models that generate Deepfakes faster.
        \item The authors should consider possible harms that could arise when the technology is being used as intended and functioning correctly, harms that could arise when the technology is being used as intended but gives incorrect results, and harms following from (intentional or unintentional) misuse of the technology.
        \item If there are negative societal impacts, the authors could also discuss possible mitigation strategies (e.g., gated release of models, providing defenses in addition to attacks, mechanisms for monitoring misuse, mechanisms to monitor how a system learns from feedback over time, improving the efficiency and accessibility of ML).
    \end{itemize}
    
\item {\bf Safeguards}
    \item[] Question: Does the paper describe safeguards that have been put in place for responsible release of data or models that have a high risk for misuse (e.g., pre-trained language models, image generators, or scraped datasets)?
    \item[] Answer: \answerNA{} 
    \item[] Justification: This work does not release pre-trained models or new datasets that pose risks for misuse.
    \item[] Guidelines:
    \begin{itemize}
        \item The answer \answerNA{} means that the paper poses no such risks.
        \item Released models that have a high risk for misuse or dual-use should be released with necessary safeguards to allow for controlled use of the model, for example by requiring that users adhere to usage guidelines or restrictions to access the model or implementing safety filters. 
        \item Datasets that have been scraped from the Internet could pose safety risks. The authors should describe how they avoided releasing unsafe images.
        \item We recognize that providing effective safeguards is challenging, and many papers do not require this, but we encourage authors to take this into account and make a best faith effort.
    \end{itemize}

\item {\bf Licenses for existing assets}
    \item[] Question: Are the creators or original owners of assets (e.g., code, data, models), used in the paper, properly credited and are the license and terms of use explicitly mentioned and properly respected?
    \item[] Answer: \answerYes{}{} 
    \item[] Justification: All datasets used are publicly available and properly cited. CIFAR-100~\cite{krizhevsky_2009_cifar100}, ImageNet-1K~\cite{deng_2009_imagenet}, CREMA-D~\cite{cao_2014_crema}, Sarcasm~\cite{cai_2019_sarcasm_dataset}, and Kinetics-400~\cite{kay_2017_kinetics} are standard research benchmarks. Baseline implementations were obtained from the authors' publicly released code repositories.
    \item[] Guidelines:
    \begin{itemize}
        \item The answer \answerNA{} means that the paper does not use existing assets.
        \item The authors should cite the original paper that produced the code package or dataset.
        \item The authors should state which version of the asset is used and, if possible, include a URL.
        \item The name of the license (e.g., CC-BY 4.0) should be included for each asset.
        \item For scraped data from a particular source (e.g., website), the copyright and terms of service of that source should be provided.
        \item If assets are released, the license, copyright information, and terms of use in the package should be provided. For popular datasets, \url{paperswithcode.com/datasets} has curated licenses for some datasets. Their licensing guide can help determine the license of a dataset.
        \item For existing datasets that are re-packaged, both the original license and the license of the derived asset (if it has changed) should be provided.
        \item If this information is not available online, the authors are encouraged to reach out to the asset's creators.
    \end{itemize}

\item {\bf New assets}
    \item[] Question: Are new assets introduced in the paper well documented and is the documentation provided alongside the assets?
    \item[] Answer: \answerNA{} 
    \item[] Justification: This paper does not release new datasets or pre-trained models. Code is provided as supplementary material for reproducibility.
    \item[] Guidelines:
    \begin{itemize}
        \item The answer \answerNA{} means that the paper does not release new assets.
        \item Researchers should communicate the details of the dataset\slash code\slash model as part of their submissions via structured templates. This includes details about training, license, limitations, etc. 
        \item The paper should discuss whether and how consent was obtained from people whose asset is used.
        \item At submission time, remember to anonymize your assets (if applicable). You can either create an anonymized URL or include an anonymized zip file.
    \end{itemize}

\item {\bf Crowdsourcing and research with human subjects}
    \item[] Question: For crowdsourcing experiments and research with human subjects, does the paper include the full text of instructions given to participants and screenshots, if applicable, as well as details about compensation (if any)? 
    \item[] Answer: \answerNA{} 
    \item[] Justification: This paper does not involve crowdsourcing or research with human subjects.
    \item[] Guidelines:
    \begin{itemize}
        \item The answer \answerNA{} means that the paper does not involve crowdsourcing nor research with human subjects.
        \item Including this information in the supplemental material is fine, but if the main contribution of the paper involves human subjects, then as much detail as possible should be included in the main paper. 
        \item According to the NeurIPS Code of Ethics, workers involved in data collection, curation, or other labor should be paid at least the minimum wage in the country of the data collector. 
    \end{itemize}

\item {\bf Institutional review board (IRB) approvals or equivalent for research with human subjects}
    \item[] Question: Does the paper describe potential risks incurred by study participants, whether such risks were disclosed to the subjects, and whether Institutional Review Board (IRB) approvals (or an equivalent approval/review based on the requirements of your country or institution) were obtained?
    \item[] Answer: \answerNA{} 
    \item[] Justification: This paper does not involve research with human subjects and therefore does not require IRB approval.
    \item[] Guidelines:
    \begin{itemize}
        \item The answer \answerNA{} means that the paper does not involve crowdsourcing nor research with human subjects.
        \item Depending on the country in which research is conducted, IRB approval (or equivalent) may be required for any human subjects research. If you obtained IRB approval, you should clearly state this in the paper. 
        \item We recognize that the procedures for this may vary significantly between institutions and locations, and we expect authors to adhere to the NeurIPS Code of Ethics and the guidelines for their institution. 
        \item For initial submissions, do not include any information that would break anonymity (if applicable), such as the institution conducting the review.
    \end{itemize}

\item {\bf Declaration of LLM usage}
    \item[] Question: Does the paper describe the usage of LLMs if it is an important, original, or non-standard component of the core methods in this research? Note that if the LLM is used only for writing, editing, or formatting purposes and does \emph{not} impact the core methodology, scientific rigor, or originality of the research, declaration is not required.
    \item[] Answer: \answerNA{} 
    \item[] Justification: The core methods in this research do not involve LLMs. All experiments were developed based on existing codebases or well-known results.
    \item[] Guidelines:
    \begin{itemize}
        \item The answer \answerNA{} means that the core method development in this research does not involve LLMs as any important, original, or non-standard components.
        \item Please refer to our LLM policy in the NeurIPS handbook for what should or should not be described.
    \end{itemize}

\end{enumerate}